\documentclass[10pt,twocolumn,letterpaper]{article}

\usepackage[margin=0.9in]{geometry} %TODO <---- remove for CVPR camera ready --- for ArXiV only!!!
\usepackage{titlesec}
\titleformat{\section}
  {\normalfont\fontsize{12.5}{15}\bfseries}{\thesection.}{0.2em}{}
\usepackage{times}
\usepackage{epsfig}
\usepackage{graphicx}
\usepackage{amsmath}
\usepackage{amssymb}
\usepackage{amsthm}
\usepackage{nicefrac}
\usepackage{algorithm}% http://ctan.org/pkg/algorithm
\usepackage{algpseudocode}% http://ctan.org/pkg/algorithmicx
\usepackage{siunitx}
\DeclareSIUnit\nsec{\text{nanoseconds}}
    \DeclareSIUnit[number-unit-product = \ ]\nanosec{\text{nanoseconds}}
\algnewcommand{\LineComment}[1]{\State \(\triangleright\) #1}

\newcommand{\lam}{\Phi_\text{bkg}}

\theoremstyle{theorem}\newtheorem{thm}{Result}
\theoremstyle{definition}\newtheorem{defn}{Definition}
\theoremstyle{remark}

\newenvironment{myenum}
{ \begin{enumerate}
    \setlength{\itemsep}{0pt}
    \setlength{\parskip}{0pt}
    \setlength{\parsep}{0pt}     }
{ \end{enumerate}                } 

\def\etal{\emph{et al}.}
\makeatother
\def\E{\mathbb E}
\def\P{\mathbb P}
\def\prob{\mathsf{P}}
\DeclareMathOperator*{\argmax}{arg\,max}
\DeclareMathOperator*{\argmin}{arg\,min}
\DeclareMathOperator{\EX}{\mathbb{E}}% expected value
\newcommand*{\defeq}{\mathrel{\vcenter{\baselineskip0.5ex \lineskiplimit0pt
                     \hbox{\scriptsize}\hbox{\scriptsize}}}%
                     =}

\usepackage[pagebackref=true,breaklinks=true,colorlinks,bookmarks=false]{hyperref}

\def\cvprPaperID{1624} % *** Enter the CVPR Paper ID here

\begin{document}
\date{}
%%%%%%%%% TITLE
\newcommand{\mytitle}{\vspace{-0.3in}Photon-Flooded Single-Photon 3D Cameras\vspace{-0.1in}}
\title{\mytitle}

%Single Photon 3D Imaging in Bright Sunlight \\ Intriguing Properties of Single Photon 3D Cameras: \\Can Single Photon LiDAR Work in Bright Sunlight? 

% \author{Anant Gupta\thanks{Department of Computer Sciences (corresponding author)}\\
% {\tt\small anant@cs.wisc.edu}\\
% \and
% Atul Ingle\thanks{Department of Computer Sciences and Department of Biostatistics}\\
% {\tt\small ingle@uwalumni.com}\\
% \and
% Andreas Velten\thanks{Department of Biostatistics and Department of Electrical and Computer Engineering}\\
% {\tt\small velten@wisc.edu}
% \and
% Mohit Gupta\thanks{Department of Computer Sciences, University of Wisconsin-Madison}\\
% {\tt\small mohitg@cs.wisc.edu}\\
% }

\author{
Anant Gupta \,\,\,
Atul Ingle \,\,\,
Andreas Velten \,\,\,
Mohit Gupta \\
{\tt \small \{agupta225,ingle,velten,mgupta37\}@wisc.edu} \vspace{3pt}\\
University of Wisconsin-Madison
}
%\author{Anant Gupta\\
%{\tt\small anant@cs.wisc.edu}\\
%{University of Wisconsin-Madison}
%\and
%Atul Ingle\\
%{\tt\small ingle@uwalumni.com}\\
%{University of Wisconsin-Madison}
%\and
%Andreas Velten\\
%{\tt\small velten@wisc.edu}\\
%{University of Wisconsin-Madison}
%\and
%Mohit Gupta\\
%{\tt\small mohitg@cs.wisc.edu}\\
%{University of Wisconsin-Madison}
%}
\maketitle
\renewcommand*{\thefootnote}{$\dagger$}
\setcounter{footnote}{1}
\footnotetext{This research was supported in part by ONR grants N00014-15-1-2652 and
N00014-16-1-2995 and DARPA grant HR0011-16-C-0025.}
\renewcommand*{\thefootnote}{\arabic{footnote}}
\setcounter{footnote}{0}
\thispagestyle{empty}
%\thispagestyle{empty}

%%%%%%%%% ABSTRACT
\begin{abstract}
Single-photon avalanche diodes (SPADs) are starting to play a pivotal role in
the development of photon-efficient, long-range LiDAR systems. However, due to
non-linearities in their image formation model, a high photon flux (e.g., due to
strong sunlight) leads to distortion of the incident temporal waveform, and
potentially, large depth errors. Operating SPADs in low flux regimes can
mitigate these distortions, but, often requires attenuating the signal and thus,
results in low signal-to-noise ratio. In this paper, we
address the following basic question: what is the optimal photon flux that a
SPAD-based LiDAR should be operated in? We derive a closed form expression for
the optimal flux, which is quasi-depth-invariant, and depends on the ambient
light strength. The optimal flux is lower than what a SPAD typically measures
in real world scenarios, but surprisingly, considerably higher than what is
conventionally suggested for avoiding distortions. We propose a simple,
adaptive approach for achieving the optimal flux by attenuating incident flux
based on an estimate of ambient light strength. Using extensive simulations and
a hardware prototype, we show that the optimal flux criterion holds for several
depth estimators, under a wide range of illumination conditions. 
\end{abstract}

%These results can enable SPAD-based LiDAR in challenging outdoor conditions with strong ambient light.

% We formalize the optimality in terms of a geometric/mathematical construct
% called bin receptivity curve.

%%%%%%%%% BODY TEXT
\vspace{-0.1in} 
\section{Introduction}
Single-photon avalanche diodes (SPAD) are increasingly being used in active
vision applications such as fluorescence lifetime-imaging microscopy (FLIM)
\cite{Shepard:2008:FLIM}, non-line-of-sight (NLOS) imaging~\cite{OToole2018},
and transient imaging \cite{OToole:2017:SPAD}. Due to their extreme sensitivity
and timing resolution, these sensors can play an enabling role in demanding
imaging scenarios, for instance, long-range LiDAR \cite{Charbon:2013:SPAD} for
automotive applications \cite{NatPhot2018}, with only limited power
budgets~\cite{Pacala:2018:patent}. 

\begin{figure}
\centering \includegraphics[width=0.9\columnwidth]{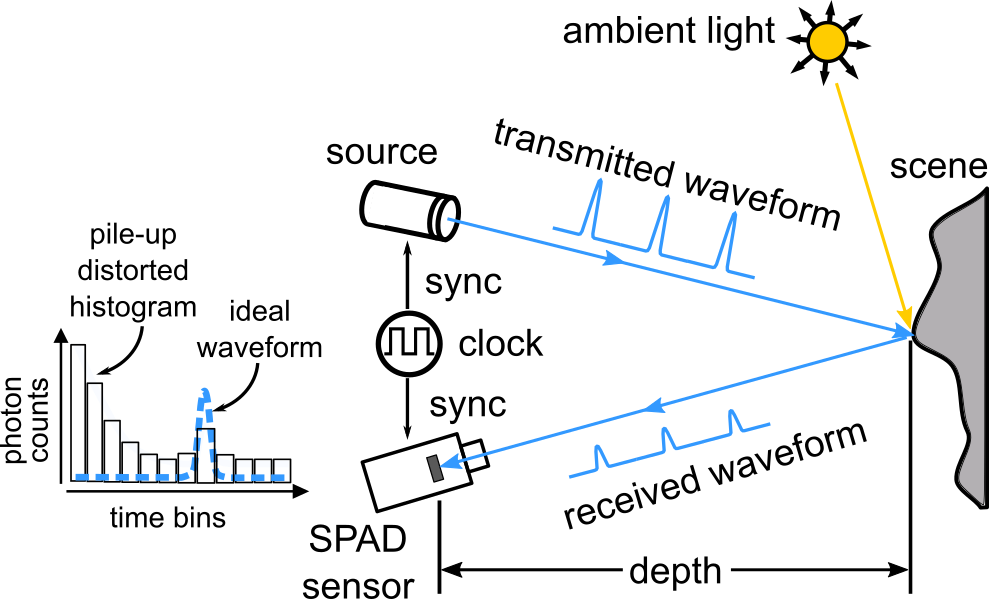}
  \caption{{\bf Pile-up in SPAD-based pulsed LiDAR.} A pulsed LiDAR consists of a light source that illuminates scene points with periodic short pulses. A SPAD sensor records the arrival times of returning photons with respect to the most recent light pulse, and uses those to build a timing histogram. In low ambient light, the histogram is the same shape as the temporal waveform received at the SPAD, and can be used for accurate depth estimation. However, in high ambient light, the histogram is distorted due to pile-up, resulting in potentially large depth errors. \label{fig:pulsed_lidar}}
  \vspace{-0.15in}
\end{figure}

A SPAD-based LiDAR (Fig.~\ref{fig:pulsed_lidar}) typically consists of a
laser which sends out periodic light pulses. The SPAD detects the first
incident photon in each laser period, after which it enters a \emph{dead time},
during which it cannot detect any further photons. The first photon detections
in each period are then used to create a histogram (over several periods) of
the time-of-arrival of the photons. If the incident flux level is sufficiently
low, the histogram is approximately a scaled version of the received temporal
waveform, and thus, can be used to estimate scene depths and reflectivity. 

Although SPAD-based LiDARs hold considerable promise due to their single-photon
sensitivity and extremely high timing (hence, depth) resolution, the peculiar
histogram formation procedure causes severe non-linear distortions due to 
% the presence of?
ambient
light \cite{Harris:1979:pileup}. This is because of intriguing characteristics
of SPADs under high incident flux: the detection of a photon depends on 
% this is not exactly correct; in TCSPC mode it depends on the arrival of photons in the same cycle, not on the time, not on photon arrivals in the previous cycle
the time of arrival of previous photons. This leads to non-linearities in the
image formation model; the measured histogram gets skewed towards earlier time
bins, as illustrated in Figs.~\ref{fig:pulsed_lidar}
and~\ref{fig:timing_diagram}\footnote{In contrast, for a conventional, linear-mode LiDAR
pixel, the detection of a photon is independent of previous photons (except
past saturation). Therefore, ambient
light adds a constant value to the entire waveform.}. This distortion, also called
``pile-up''~\cite{Harris:1979:pileup}, becomes increasingly severe as the
amount of ambient light increases, and can lead to large depth errors. This can
severely limit the performance of SPAD-based LiDAR in outdoor conditions, for
example, imagine a power-constrained automotive LiDAR operating on a bright
sunny day~\cite{NatPhot2018}. 

One way to mitigate these distortions is to attenuate the incident flux
sufficiently so that the image formation model becomes approximately
linear~\cite{patting2018fluorescence,Heide:2018:subpicosecond}. However, in a
LiDAR application, most of the incident flux may be due to ambient light. In
this case, lowering the flux (e.g., by reducing aperture size), requires
attenuating \emph{both} the ambient and the signal light\footnote{Ambient
light can be reduced to a limited extent via spectral filtering.}. While this
mitigates distortions, it also leads to signal loss. This fundamental tradeoff
between distortion (at high flux) and low signal (at low flux) raises a natural
question: Is there an optimal incident flux for SPAD-based active 3D imaging
systems? \smallskip

\noindent{\bf Optimal incident flux for SPAD-based LIDAR:} We address this
question by analyzing the non-linear imaging model of SPAD LiDAR. Given
a fixed ratio of source-to-ambient light strengths, we derive a closed-form
expression for the optimal incident flux. Under certain assumptions, the
optimal flux is quasi-invariant to source strength and scene depths, and
surprisingly, depends only on the ambient strength and the unambiguous depth
range of the system. Furthermore, the optimal flux is lower than that
encountered by LiDARs in typical outdoor conditions. This suggests
that, somewhat counter-intuitively, reducing the total flux improves
performance, even if that means attenuating the signal. On the other hand,
the optimal flux is considerably higher than that needed for the image
formation to be in the linear
regime~\cite{becker2015advanced,Kapusta:2015:TCSPC}. As a result, while the
optimal flux still results in some degree of distortion, with appropriate
computational depth estimators, it achieves high performance across
a wide range of imaging scenarios.  

Based on this theoretical result, we develop a simple adaptive scheme for SPAD
LiDAR where the incident flux is adapted based on an estimate of the ambient
light strength. We perform extensive simulation and hardware experiments to
demonstrate that the proposed approach achieves up to an order of magnitude
higher depth precision as compared to existing \emph{rule-of-thumb}
approaches~\cite{becker2015advanced,Kapusta:2015:TCSPC} that require lowering
flux levels to linear regimes. \smallskip

\noindent{\bf Implications:} The theoretical results derived in this paper can
lead to a better understanding of this novel and exciting sensing technology.
Although our analysis is performed for an analytical pixel-wise depth
estimator~\cite{Coates}, we show that in practice, the improvements in depth
estimation are achieved for several reconstruction approaches, including
pixel-wise statistical approaches such as MAP, as well as estimators that
account for spatial correlations and scene priors (e.g., neural network
estimators~\cite{Lindell}). These results may motivate the design of practical,
low-power LiDAR systems that can work in a wide range of illumination
conditions, ranging from dark to extreme sunlight.

\section{Related Work}

\noindent{\bf SPAD-based active vision systems:} Most SPAD-based LiDAR, FLIM
and NLOS imaging systems
\cite{Buttafava:15,Kirmani58,Shin,Rapp,Lindell,ambient_rejection} rely on the
incident flux being sufficiently low so that pile-up distortions can be
ignored. Recent work~\cite{Heide:2018:subpicosecond}
has addressed the problem of source light
pile-up for SPAD-based LiDAR using a realistic model of the
laser pulse shape and statistical priors on scene structure to
achieve sub-pulse-width depth precision.
% Edited for Arxiv
%This work uses a realistic model of the laser pulse shape and statistical priors to provide 
%sub-picosecond accuracy in situations where signal light is the dominant source of flux.
% This work models pile-up only due to non-impulse laser pulses, and assumes
% negligible/low ambient light.
Our goal is different---we provide theoretical analysis and
design of SPAD LiDAR that can perform robustly even in strong \emph{ambient} light. \smallskip
% are we no longer talking about methods which rely on scene correlations and frame a joint optimization problem over the entire depth map? eg. Rapp and Goyal, neural net and sub-pico? 

\noindent{\bf Theoretical analysis and computational methods for pile-up
correction:} Pile-up distortion can be removed in post-processing by
computationally inverting the non-linear image formation
model~\cite{Coates,Walker:2002:pileup}. While these approaches can mitigate
relatively low amount of pile-up, they have limited success in high flux
levels, where a computational approach alone results in strong amplification of
noise. Previous work has performed theoretical analysis similar to ours in a
range-gating scenario where scene depths are
known~\cite{fouche2003detection,wang2018adaptive,degnan2008impact}. In
contrast, we derive an optimal flux criterion that minimizes pile-up errors at
capture time, is applicable for a broad range of, including extremely high,
lighting levels, and does not require prior knowledge of scene depths.
\smallskip

\noindent{\bf Alternative sensor architectures:} Pile-up can be suppressed by
modifying the detector hardware, eg. by using multiple SPADs per
pixel connected to a single time-correlated single-photon counting (TCSPC)
circuit to distribute the high incident flux
over multiple SPADs \cite{ambient_rejection}.  Multi-SPAD schemes with parallel
timing units and multi-photon thresholds can be used to detect correlated
signal photons \cite{Perenzoni2017A6} and reject ambient light photons that are
temporally randomly distributed. The theoretical criteria derived here can be
used in conjunction with these hardware architectures for optimal LiDAR design.
\smallskip

\noindent{\bf Active 3D imaging in sunlight:} Prior work in the structured
light and time-of-flight literature proposes various coding and illumination
schemes to address the problem of low signal-to-noise ratios (SNR) due to
strong ambient light \cite{mertz2012low,structured,episcan,epitof}. The present work deals with a different problem of
optimal photon detection for SPAD-based pulsed time-of-flight. These previous
strategies can potentially be applied in combination with our method to further
improve depth estimation performance.

\section{Background: SPAD LiDAR Imaging Model}
\begin{figure*}[!ht]
\centering \includegraphics[width=0.9\textwidth]{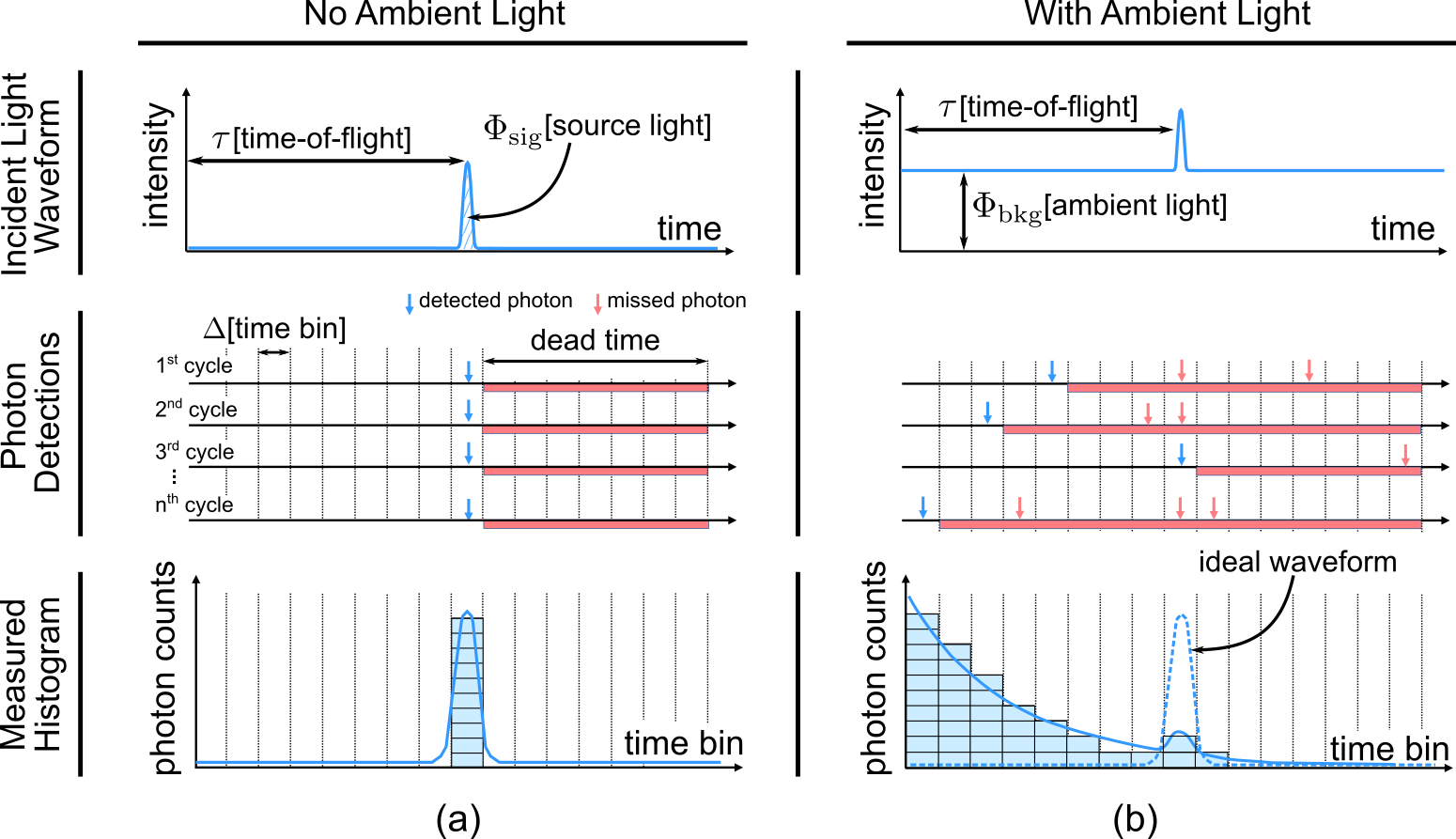}
   \caption{{\bf Effect of ambient light on SPAD LiDAR.} A SPAD-based pulsed LiDAR builds a histogram of the time-of-arrival of the incident photons, over multiple laser pulse cycles. In each cycle, at most one photon is recorded, whose timestamp is used to increment the counts in the corresponding histogram bin. (Left) When there is no ambient light, the histogram is simply a discretized, scaled version of the incident light waveform. (Right) Ambient light photons arriving before the laser pulse skew the shape of the histogram, causing a non-linear distortion, called pile-up. This results in large depth errors, especially as ambient light increases. 
   \label{fig:timing_diagram}}
   \vspace{-0.1in}
\end{figure*}

This section provides mathematical background on the 
image formation model for SPAD-based pulsed LiDAR.
Such a system typically consists of a laser source which
transmits periodic short pulses of light at a scene point, and a co-located
SPAD detector \cite{oconnor,RENKER200648,Dautet:93} which observes the
reflected light, as shown in Fig.~\ref{fig:pulsed_lidar}. We model an ideal
laser pulse as a Dirac delta function $\tilde \delta(t)$. Let $d$ be the
distance of the scene point from the sensor, and $\tilde\tau  =
\nicefrac{2d}{c}$ be the round trip time-of-flight for the light pulse. The photon flux
incident on the SPAD is given by: 
\setlength{\abovedisplayskip}{4pt}
\setlength{\belowdisplayskip}{4pt}
\begin{equation}
  \Phi(t) = \tilde \Phi_\text{sig} \; \tilde \delta(t - \tilde \tau) + \tilde \Phi_\text{bkg},
  \label{eq:waveform}
\end{equation}
where $\tilde \Phi_\text{sig}$ is the signal component of the received
waveform; it encapsulates the laser source power, distance-squared fall-off, scene
brightness and BRDF. $\tilde \Phi_\text{bkg}$ denotes the background component,
assumed to be a constant due to ambient light. 
% constant over a time scale of few hundreds of nanoseconds, which is the typical acquisition time
Since SPADs have a finite time resolution (few tens of picoseconds), we
consider a discretized version of the continuous waveform in
Eq.~(\ref{eq:waveform}), using uniformly spaced time bins of size $\Delta$. Let
$M_i$ be the number of photons incident on the SPAD in the $i^\text{th}$ time
bin. Due to arrival statistics of photons, $M_i$ follows a Poisson
distribution. The mean of the Poisson distribution, $\EX[M_i]$, i.e., the
average number $r_i$ of photons incident in $i^\text{th}$ bin, is given as:
\begin{align}
  r_i &= \Phi_\text{sig} \; \delta_{i,\tau} + \Phi_\text{bkg}. \label{eq:poisson_rates}
\end{align}
Here, $\delta_{i,j}$ is the Kronecker delta,\footnote{The
Kronecker delta is defined as $\delta_{i,j}=1$ for $i=j$ and $0$ otherwise.}
$\Phi_\text{sig}$ is the mean number of signal photons received per bin, 
and $\Phi_\text{bkg}$ is the (undesirable) background and dark count photon flux per bin. Let $B$ be the total number of time bins. Then, we define the vector of values $\left(r_1, r_2, \ldots, r_B
\right)$ as the \emph{ideal} incident waveform. \smallskip

% should we have a couple of lines explaining why the TCSPC mode is generally used, as opposed to free running?
\noindent{\bf SPAD histogram formation:} SPAD-based LiDAR systems operate on
the TCSPC principle 
\cite{Kapusta:2015:TCSPC}. A scene point is illuminated by a periodic train
of laser pulses. Each period starting with a laser pulse is referred to as a
\emph{cycle}. The SPAD detects only the first incident photon in each cycle,
after which it enters a dead time (${\sim}\SI{100}{\nano\second}$),
during which it cannot detect any further photons. The time of arrival of the
first photon is recorded with respect to the start of the most recent
cycle. A histogram of first photon arrival times is constructed over many laser
cycles, as shown in Fig.~\ref{fig:timing_diagram}.

If the histogram consists of $B$ time bins, the laser repetition period is
$B\Delta$, corresponding to an unambiguous depth range of
$d_\text{max}=cB\Delta/2$.  Since the SPAD only records the first photon in
each cycle, a photon is detected in the $i^\text{th}$ bin only if at least one
photon is incident on the SPAD during the $i^\text{th}$ bin, and, no photons
are incident in the preceding bins. The probability $q_i$ that at least one
photon is \emph{incident} during the $i^\text{th}$ bin can be computed using
the Poisson distribution with mean $r_i$ \cite{Coates}: 
\[
  q_i = \prob(M_i \geq 1) = 1-e^{-r_i}.
\]
\noindent Thus, the probability $p_i$ of \emph{detecting} a photon in the
$i^\text{th}$ bin, in any laser cycle, is given by \cite{pediredla2018signal}:
\begin{align}
p_i &= q_i \; \prod_{k=1}^{i-1} (1-q_k) = \left(1 - e^{-r_i}\right) e^{-\sum_{k = 1}^{i - 1} r_k}.
\label{eq:prob_photon_detection}
\end{align}
Let $N$ be the total number of laser cycles used for forming a histogram and
$N_i$ be the number of photons detected in the $i^\text{th}$ histogram bin. The
vector $(N_1,\!N_2,\!\ldots,\!N_{B+1})$ of the histogram counts follows a
multinomial distribution: 
\begin{equation}
  (N_1,\!N_2,\!\ldots,\!N_{B+1}) \!\sim \!\mathsf{Mult}(N,(p_1,p_2,\ldots,p_{B+1}))\,,
  \label{eq:histogram_model}
\end{equation}
where, for convenience, we have introduced an additional $(B+1)^\text{st}$
index in the histogram to record the number of cycles with no detected photons.
Note that $p_{B+1} \defeq 1-\sum_{i=1}^{B}p_i$ and $N=\sum_{i=1}^{B+1}N_i.$
Eq.~(\ref{eq:histogram_model}) describes a general probabilistic model for the
histogram of photon counts acquired by a SPAD-based pulsed LiDAR.

Fig.~\ref{fig:timing_diagram} (a) shows the histogram formation in the case of
negligible ambient light. In this case, all the photon arrival times line up
with the location of the peak of the incident waveform. As a result, $r_i=0$
for all the bins except that corresponding to the laser impulse
peak. In this case, the measured histogram vector
$(N_1,\!N_2,\!\ldots,\!N_{B})$, on average, is simply a scaled version of the
incident waveform $\left(r_1, r_2, \ldots, r_B \right)$. The time-of-flight can
be estimated by locating the bin index with the highest photon counts:
\begin{equation}
  \widehat \tau = \argmax_{1\leq i \leq B}\, N_i \,, \label{eq:time_of_arrival_estimator}
\end{equation}
and the scene depth can be estimated as $\widehat d = \frac{c \widehat \tau \Delta }{2}$.

% Edited for Arxiv
For ease of theoretical analysis, we assume the laser pulse is a perfect Dirac-impulse with a duration of a single
time bin. We also ignore other SPAD non-idealities such as jitter and
afterpulsing.
% We use a simplified Dirac-impulse for ease of theoretical analysis.
We show in the supplement that the results presented here can potentially be improved by combining our optimal
photon flux criterion with recent work \cite{Heide:2018:subpicosecond}
that explicitly models the laser pulse shape and SPAD timing jitter.
\section{Effect of Ambient Light on SPAD LiDAR} 
If there is ambient light, the waveform incident on the SPAD can be modeled as
an impulse with a constant vertical shift, as shown in the top of
Fig.~\ref{fig:timing_diagram} (b). The measured histogram, however, does not
reliably reproduce this ``DC shift'' due to the peculiar histogram formation
procedure that only captures the first photon for each laser cycle. When the
ambient flux is high, the SPAD detects an ambient photon in the earlier
histogram bins with high probability, resulting in a distortion with an
exponentially decaying shape. This is illustrated in the bottom of
Fig.~\ref{fig:timing_diagram} (b), where the peak due to laser source appears
only as a small blip in the exponentially decaying tail of the measured
histogram. The problem is exacerbated for scene points that are farther from
the imaging system. This distortion, called pile-up, significantly lowers the
accuracy of depth estimates because the bin corresponding to the true depth no
longer receives the maximum number of photons. In the extreme case, the later
histogram bins might receive no photons, making depth reconstruction at those
bins impossible.
\smallskip

\begin{figure}
\centering \includegraphics[width=1.0\columnwidth]{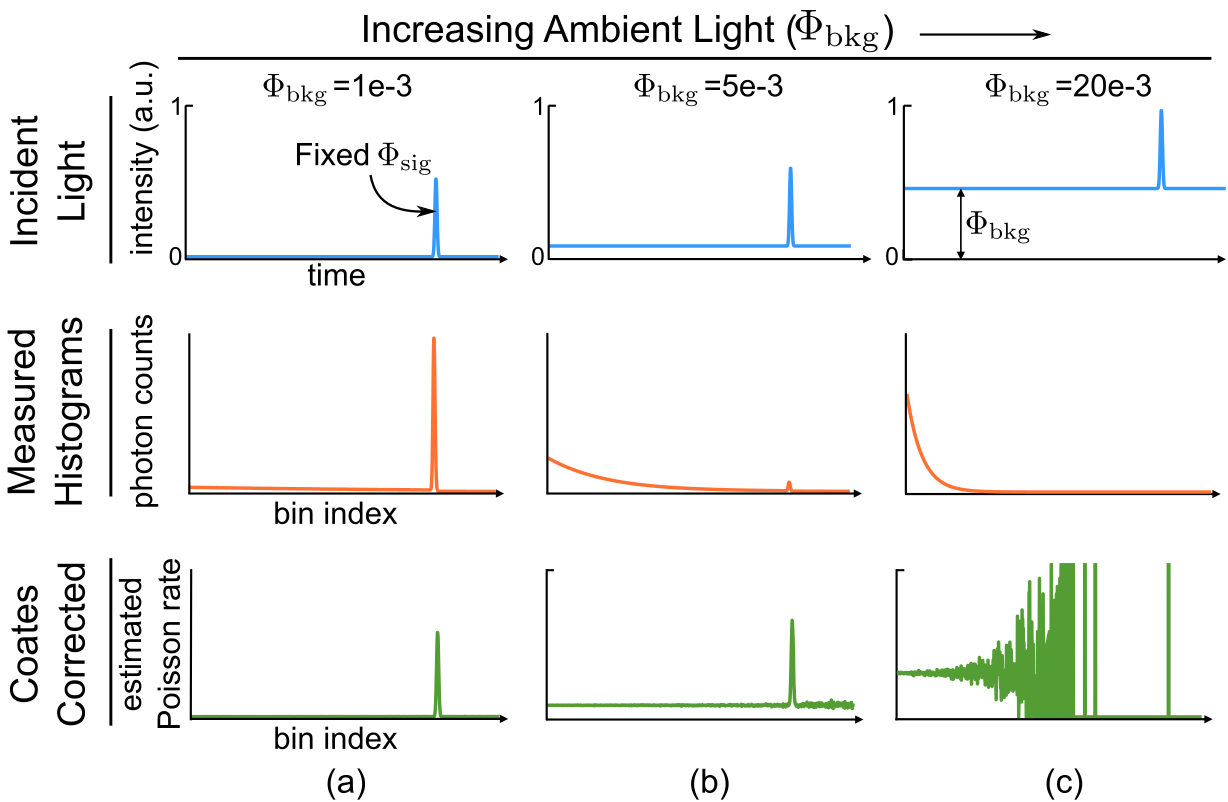}
  \caption{{\bf Efficacy of computational pile-up correction approaches~\cite{Coates}.}
  (a) In low ambient light, there is negligible pile-up. (b) At moderate ambient
  light levels, pile-up can be observed as a characteristic exponential
  fall-off in the acquired histogram. The signal pulse location can still be
  recovered using computational correction (Section~\ref{sec:Coates}). (c) In
  strong ambient light, the later histogram bins receive very few photons,
  which makes the computationally corrected waveform extremely noisy, making it
  challenging to reliably locate the laser peak for estimating depth.
  \label{fig:coates_correction}}
  \vspace{-0.1in}
\end{figure}
% should we explain how the Coates' corrected rates can be used to estimate depth by taking the argmax, or is it obvious?

\noindent{\bf Computational Pile-up Correction:}\label{sec:Coates}
In theory, it is possible to ``undo'' the distortion by inverting the
exponential nonlinearity of Eq.~(\ref{eq:prob_photon_detection}), and finding
an estimate of the incident waveform $r_i$ in terms of the measured histogram
$N_i$:
\begin{equation}
  \widehat{r}_i = \ln{\left(\frac{N - \sum_{k=1}^{i-1} N_k}{N - \sum_{k=1}^{i-1} N_k - N_i}\right)}.
  \label{eq:coates_corrected_rate_estimate}
\end{equation}
This method is called the Coates's correction~\cite{Coates}, and it can be
shown to be equivalent to the maximum-likelihood estimate of $r_i$
\cite{pediredla2018signal}. See supplementary document for a self-contained
proof. The depth can then be estimated as:
\begin{equation}
\widehat \tau = \argmax_{1\leq i \leq B}{\widehat r_i}.
\label{eq:coates_estimator}
\end{equation}
Although this computational approach removes distortion, the
non-linear mapping from measurements $N_i$ to the estimate $\widehat{r}_i$
significantly amplifies measurement noise at later time bins, as shown in Fig.~\ref{fig:coates_correction}. \smallskip
% Figs.~\ref{fig:coates_correction}~(a)~and~(b) show that although Coates's
% correction successfully extracts the laser pulse peak buried in the tail of the
% exponentially decaying histogram for low and moderate ambient light, it fails
% at high ambient light, resulting in high variance at later bins, as shown in
% Fig.~\ref{fig:coates_correction}~(c).  

\noindent {\bf Pile-up vs. Low Signal Tradeoff:}
One way to mitigate pile up is to reduce the total incident
photon flux (e.g., by reducing the aperture or SPAD size). Various
rules-of-thumb \cite{becker2015advanced,Kapusta:2015:TCSPC} advocate
maintaining a low enough photon flux so that only 1-5\% of the laser cycles
result in a photon being detected by the SPAD.
%\footnote{\scriptsize For example, consider a 5 MHz laser
%(\SI{200}{\nano\second} repetition period). The 1-5\% rule suggests that the
%flux should be 50,000-250,000 photons/second. Assuming a histogram with
%$B\!=\!2000$ bins, we have $\Delta=\SI{100}{\pico\second}.$ So the mean
%Poisson rates $r_i$ should be between $5\!\times \!10^{-6}$ and $25\!\times
%\!10^{-6}$ photons per bin.}
In this case, $r_i \ll 1 \, \forall \, i$ and Eq.~(\ref{eq:prob_photon_detection}) simplifies to $p_i \approx
r_i$. Therefore, the mean photon counts $N_i$ become proportional to the
incident waveform $r_i$, i.e., $\E[N_i] = Np_i \approx Nr_i.$
This is called the \emph{linear operation regime} because the measured
histogram $(N_i)_{i=1}^B$ is, on average, simply a scaled version of the true
incident waveform $(r_i)_{i=1}^B.$  This is similar to the case of no ambient
light as discussed above, where depths can be estimated by locating the
histogram bin with the highest photon counts. 

Although lowering the overall photon flux to operate in the linear regime
reduces ambient light and prevents pile-up distortion, unfortunately, it also
reduces the source signal considerably. On the other hand, if the incident
photon flux is allowed to remain high, the histogram suffers from pile-up,
undoing which leads to amplification of noise.  This fundamental tradeoff
between pile-up distortion and low signal raises a natural question: What is
the optimal incident flux level for the problem of depth estimation using
SPADs?

\section{Bin Receptivity and Optimal Flux Criterion}
In this section, we formalize the notion of optimal incident photon flux for a
SPAD-based LiDAR. We model the \emph{original} incident waveform as a constant
ambient light level $\Phi_\text{bkg}$, with a single source light pulse of
height $\Phi_\text{sig}$. We assume that we can modify the incident waveform
only by attenuating it with a scale factor $\Upsilon \leq 1$. This attenuates
\emph{both} the ambient $\Phi_\text{bkg}$ and source $\Phi_\text{sig}$
components proportionately.~\footnote{It is possible to selectively attenuate
only the ambient component, to a limited extent, via spectral filtering. We
assume that the ambient level $\Phi_\text{bkg}$ is already at the minimum
level that is achievable by spectral filtering.} Then, given a
$\Phi_\text{bkg}$ and $\Phi_\text{sig}$, the total photon flux incident on the
SPAD is determined by the factor $\Upsilon$. Therefore, the problem of finding
the optimal total incident flux can be posed as determining the optimal
attenuation $\Upsilon$. To aid further analysis, we define the following term. 
%Anant comments
% There are 2 main approximations that we make, sig << B bkg, and bkg << 1. The
% first one is needed to make the bin receptivity curve a function of only r,
% and I think we are mentioning it at the right place, just before eqn 8. Maybe
% we can have another step showing how exactly the approximation leads to the
% equation.
%The second approximation is only really needed to get a cleaner optimal flux
%criterion (instead of e^(r/B) = B/(B-1)). So maybe we can get rid of it in eqn
%8 (and have C_i(r) = B*(1-e^(-r/B))*exp(-(i-1)r/B)), as well as in remark 5.5.
%So wherever we have r, we replace it by B*(1-e^(-r/B)), and the approximately
%equal signs will become exact. It's upto you whether to still use the
%approximation in eqn 9 to get a clean expression. If you think yes, then we
%can mention the approximation just before eqn 9.  As we talked in the meeting
%today, the theorem should have C_min directly, and all the approximations
%needed to derive it should be mentioned in supplementary.  I'm concerned that
%some reviewer might say that the reason you get an optimal flux condition
%independent of sig is because of the assumption sig << B bkg, which is true;
%if we use the exact value of C_min, we will get B * bkg + sig = 1 instead of
%B*bkg = 1 in the optimality condition. So maybe we can get rid of the first
%approximation too in the derivation, get a sig dependent optimality condition,
%and then use the approximation.

\begin{figure} 
\centering \includegraphics[width=1.0\columnwidth]{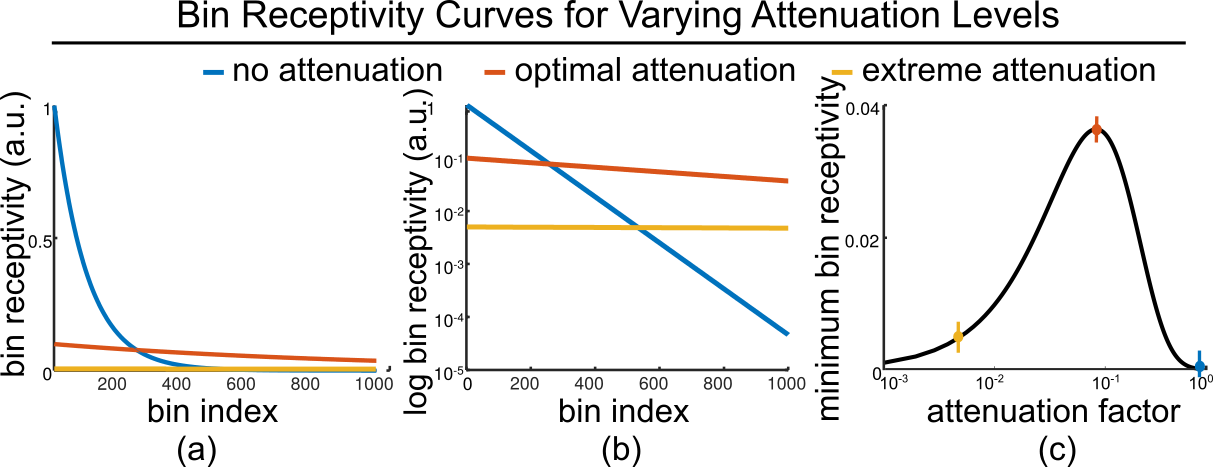} 
  \caption{ {\bf Bin receptivity curves (BRC) for different attenuation levels.} (a-b) Large (extreme) attenuation results in flat BRC with no pile-up, but low signal level. No attenuation results in a distorted BRC, but higher signal level. The proposed optimal attenuation level achieves a BRC with both low distortion, and high signal. (c) The optimal attenuation factor is given by the maxima location (unique) of the minimum value of BRC. 
	\label{fig:bin_receptivity}}
    \vspace{-0.1in}
\end{figure} 

%Ideally, we would like the detection probabilities to reflect the shape of the
%incident waveform. This is true for the

%\begin{defn}{\bf [Total Incident Flux]} The \emph{total incident flux} $r$ is
%defined as the sum of the average number of incident photons over all bins: $r
%\defeq \sum_{i=1}^B r_i$.
%\end{defn}
%Since the total incident flux includes \emph{both} the source and ambient
%photons, it should be high to maximize the SNR.

\begin{defn}{{\bf [Bin Receptivity Coefficient]}}
  The \emph{bin receptivity coefficient} $C_i$ of the $i^\text{th}$ histogram bin
  is defined as:
  \begin{equation}
      C_i \defeq \frac{p_i}{r_i} r \,, \label{eq:defC}
  \end{equation}
\end{defn}

\noindent where $p_i$ is the probability of detecting a photon   (Eq.~(\ref{eq:prob_photon_detection})), and $r_i$ is the average number of incident photons (Eq.~(\ref{eq:poisson_rates})) in the 
$i^\text{th}$ bin. $r$ is the total incident flux $r \defeq \sum_{i=1}^B r_i \,.$ The \emph{bin receptivity curve} (BRC) is defined as the plot of the bin receptivity coefficients $C_i$ as a function of the bin index $i$. 

The BRC can be considered an intuitive indicator of the performance of a SPAD LiDAR
system, since it captures the pile-up vs. shot noise tradeoff. The first term $\frac{p_i}{r_i}$ quantifies the distortion in the shape of the
measured histogram with respect to the ideal incident waveform, while the second term $r$ quantifies the strength of the signal. 
Figs.~\ref{fig:bin_receptivity} (a-b) show the BRCs for high and low incident
flux, achieved by using a high and low attenuation $\Upsilon$, respectively.
For small $\Upsilon$ (low flux), the BRC is uniform (negligible
pile-up, as $\frac{p_i}{r_i} \approx 1$ is approximately constant across $i$), but the curve's values are small (low signal). For large
$\Upsilon$ (high flux), the curve's values are large on average (large signal),
but skewed towards earlier bins (strong pile-up, as $\frac{p_i}{r_i}$ varies considerably from $\approx 1$ for earlier bins to $\ll 1$ for later bins). Higher the flux, larger the variation in $\frac{p_i}{r_i}$ over $i$. \smallskip

%A uniform bin receptivity curve (all coefficients $C_i$ approximately the
%same) is desirable since it indicates low pile up. On the other hand, a curve
%with large area under the curve (large coefficients) is also desirable since
%it indicates large signal strength, and thus, high SNR. 

\noindent {\bf BRC as a function of attenuation factor $\Upsilon$:} Assuming
total background flux $B \Phi_\text{bkg}$ over the entire laser period to be
considerably stronger than the total source flux, i.e., $\Phi_\text{sig} \ll B
\Phi_\text{bkg}$, the flux incident in the $i^\text{th}$ time bin can be
approximated as $r_i \approx \nicefrac{r}{B}$. Then, using
Eqs.~(\ref{eq:defC}) and~(\ref{eq:prob_photon_detection}), the BRC
can be expressed as:
\begin{equation}
  C_{i} = B \, (1-e^{-\frac{r}{B}}) \, e^{-(i-1) \frac{r}{B}}.
  \label{eq:bin_receptivity}
%&= \frac{r}{r_i} (1-e^{-r_i})\,e^{ -\sum_{k=1}^{i-1} r_k } \nonumber \\
%&= B \; \left[1 - \exp{\left(-\frac{r}{B}\right)}\right] \; \exp{\left(-(i-1)\frac{r}{B}\right)} \nonumber \\
\end{equation}
\noindent Since total incident flux $r = \Upsilon \, \left(\Phi_\text{sig} + B
\, \Phi_\text{bkg} \right)$, and we assume $\Phi_\text{sig} \ll B
\Phi_\text{bkg}$, $r$ can be approximated as $r \approx \Upsilon \, B \,
\Phi_\text{bkg}$. Substituting in Eq.~(\ref{eq:bin_receptivity}), we get an
expression for BRC as a function \emph{only} of the attenuation $\Upsilon$, for
a given number of bins $B$ and a background flux $\Phi_\text{bkg}$:
\begin{equation}
  C_{i} (\Upsilon) = B \, (1-e^{-\Upsilon \, \Phi_\text{bkg} }) \, e^{-(i-1) \Upsilon \, \Phi_\text{bkg}}.
  \label{eq:bin_receptivity_Upsilon}
%&= \frac{r}{r_i} (1-e^{-r_i})\,e^{ -\sum_{k=1}^{i-1} r_k } \nonumber \\
%&= B \; \left[1 - \exp{\left(-\frac{r}{B}\right)}\right] \; \exp{\left(-(i-1)\frac{r}{B}\right)} \nonumber \\
\end{equation}

Eq.~(\ref{eq:bin_receptivity_Upsilon}) allows us to navigate the space of BRCs,
and hence, the shot noise vs. pile-up tradeoff, by varying a single parameter:
the attenuation factor $\Upsilon$. Based on
Eq.~(\ref{eq:bin_receptivity_Upsilon}), we are now ready to define the optimal
$\Upsilon$. 
% that minimizes an upper bound on the mean probability of depth error ($L^0$ error) of a SPAD-LiDAR.
\begin{thm}[{\bf Attenuation and Probability of Depth Error}]
\label{res:}
Let $\tau$ be the true depth bin and $\widehat\tau$ the estimate obtained
  using the Coates's estimator
  (Eq.(\ref{eq:coates_estimator})). An upper bound on the average
  probability of depth error $\sum_{\tau=1}^B \P(\widehat\tau \neq \tau)$ is
  minimized when the attenuation fraction is given by:
  \begin{equation}
  %\underbrace{
    \boxed{
      \Upsilon^\text{\normalfont opt} = \argmax_{\Upsilon} \min_i C_i (\Upsilon).
    }
  %}%_{\text{Optimal Flux Attenuation Factor}}
  \label{eq:optimality_condition_First}
\end{equation}
\end{thm}
% \noindent {\bf Optimal $\Upsilon$:} Assuming the analytic depth
% estimator~\cite{Coates} described in Section~\ref{sec:Coates} is used for
% recovering depths, the optimal $\Upsilon$ is given in terms of the BRC as:

%In order to model the effect of total incident flux (parameterized by the
%attenuation factor $\Upsilon$) on a SPAD LiDAR's performance, next we derive
%an expression for the BRC as a function of $\Upsilon$. 
%In general, the bin receptivity vector depends on the the incident waveform,
%and good vectors with desirable properties can be designed by modifying the
%incident waveform. Let $r$ and $r_i$ are the original total and bin-wise
%average incident flux, respectively. Then, the modified total flux
%$\overline{r} = \Upsilon r$, and the modified bin-wise average flux
%$\overline{r_i} = \Upsilon r_i$.  

%\noindent {\bf BRC as a function of $\Upsilon$:} In challenging outdoor
%scenarios with strong ambient light, the  incident on the SPAD is  and thus,
%the flux incident in a time bin can be approximated as $r_i \approx
%\nicefrac{r}{B}$,. Then, using Eqs.~\ref{eq:defC},~\ref{eq:defChi}
%and~\ref{eq:prob_photon_detection}, the BRC can be expressed as a function
%only of the attenuation factor $\Upsilon$:

%\begin{thm}[]{\bf [Optimal Flux Attenuation Factor]\label{prob_error_thm}} For
%a given signal $\Phi_\text{sig}$ and background ${B \Phi_\text{bkg}}$, the
%optimal total flux is given by: \[ \Upsilon^\text{opt} = \argmax_{r} \min_i
%C_i (\Upsilon).  \] \end{thm}
See the supplementary technical report for a proof. This result states that,
given a signal and background flux, the optimal depth estimation performance is
achieved when the minimum bin receptivity coefficient is maximized. 

%\begin{thm}{\bf Probability of depth estimation error.\label{prob_error_thm}}
%  Let $\Lambda\defeq\nicefrac{\Phi_\text{sig}}{\Phi_\text{bkg}}$ be the fixed
%  signal-to-background ratio. Suppose the true depth bin $\tau$ is chosen
%  uniformly randomly from $\{1, 2, \ldots, B\}.$ Then the probability of error
%  in estimating the arrival time bin using
%  Equation~(\ref{eq:time_of_arrival_estimator}) satisfies the following upper
%  bound:
%  \[
%    \prob(\widehat \tau \neq \tau) \!\leq \!\frac{1}{B} \!\!\sum_{\substack{i,j=1\\[1pt] j\neq i}}^B\!\!
%    \exp\!\left(\!
%      -\frac{B\Lambda^2}{2(B\!+ \!\Lambda) \left( \frac{1}{C_j(r)}\!+\!(1\!+\!\Lambda)\frac{1}{C_i(r)}\right) }
%    \!\right)\!.
%  \]
%\end{thm}

%To minimize the probability of depth estimation error, intuitively, the upper
%bound given by Theorem~\ref{prob_error_thm} must be minimized. Note that the
%upper bound is dominated by the smallest exponent, i.e., the smallest bin
%receptivity value, attained at the last histogram bin $i=B$
%(Eq.~\ref{{eq:bin_receptivity}}). 
From Eq.~(\ref{eq:bin_receptivity_Upsilon}) we note that for a fixed
$\Upsilon$, the smallest receptivity value is attained at the last
bin $i=B$, i.e., $\min_i C_i (\Upsilon) = C_B (\Upsilon)$. Substituting in
Eq.~(\ref{eq:optimality_condition_First}), we get:
\[
  \Upsilon^\text{opt} = \argmax_{\Upsilon} C_B(\Upsilon). 
\]
Using $C_B(\Upsilon)$ from Eq.~(\ref{eq:bin_receptivity_Upsilon}) and solving
for $\Upsilon$, we get:
\[
  \Upsilon^\text{opt} = \frac{1}{\Phi_\text{bkg}} \, \log\left(\frac{B}{B-1}\right).
\]
%In the high ambient light case, the total flux is dominated by the background
%flux. Denoting the optimum background flux by $\Phi^\text{opt}_\text{bkg}$ and
%substituting $r^\text{opt} = B \Phi^\text{opt}_\text{bkg}$ we get:
%\[
%  \Phi^\text{opt}_\text{bkg} = \log\left(\frac{B}{B-1}\right).
%\]
Finally, assuming that $B\gg 1$, we get $\log\left(\frac{B}{B-1}\right)\approx \frac{1}{B}$.
Since $B = \nicefrac{2 d_\text{max}}{c \Delta}$, where $d_\text{max}$ is the unambiguous depth range, the final
optimality condition can be written as:
\begin{equation}
  \underbrace{
    \boxed{
      \Upsilon^\text{opt} = \frac{c \, \Delta}{2 \, d_\text{max} \, \Phi_\text{bkg}} \,. 
    }
  }_{\text{Optimal Flux Attenuation Factor}}
  \label{eq:optimality_condition}
\end{equation}

\noindent{\bf Geometric interpretation of the optimality criterion:} 
% The
% optimization objective $\min_i C_i(\Upsilon)$
% (Eq.~\ref{eq:optimality_condition_First}) can be decomposed as:
% \begin{align*}
% \min_i C_i(\Upsilon) &= C_B(\Upsilon) = B \, (1-e^{-\Upsilon \, \Phi_\text{bkg} }) \, e^{-(B-1) \Upsilon \, \Phi_\text{bkg}} \\
%               & \hspace{-0.6in} = \left[1 - e^{-\Upsilon B \Phi_\text{bkg}}\right]
%                 \frac{1}{\frac{1}{B(1-e^{-\Upsilon \Phi_\text{bkg}}) e^{(-\Upsilon B \Phi_\text{bkg})}} -
%                 \frac{1}{B(1-e^{-\Upsilon \Phi_\text{bkg}}) }} \\
%               &\hspace{-0.6in} =  \underbrace{\frac{1}{B}\sum_{i=1}^B C_i(\Upsilon)}_\text{Mean receptivity} 
%               \left(\underbrace{\frac{1}{C_{B}(\Upsilon)} - \frac{1}{C_1(\Upsilon)}}_\text{Skew}\right)^{-1}.
% \end{align*}
% The first term is the mean receptivity (area under the BRC). The second term is
% a measure of the non-uniformity (skew) of the BRC. Since the optimal $\Upsilon$
% maximizes the objective $\min_i C_i(\Upsilon)$, which is the ratio of mean
% receptivity and skew, it simultaneously achieves low distortion and large mean
% values. 
% in the figure, we should mark the minimum bin receptivities, and point out that the opt filtering one has the highest
Result~\ref{res:}
can be intuitively understood in terms of the space of shapes of the BRC.
Figs.~\ref{fig:bin_receptivity} (a-b) shows the effect of three
different attenuation levels on the BRC of a SPAD exposed to high ambient
light. When no attenuation is used, the BRC decays rapidly due to strong pile-up. Current approaches~\cite{becker2015advanced,Kapusta:2015:TCSPC} that use
\emph{extreme attenuation}~\footnote{For example, consider a depth
range of $\SI{100}{\meter}$ and a bin resolution of
$\Delta=\SI{100}{\pico\second}$. Then, the 1\% rule of thumb recommends
extreme attenuation so that each bin receives $\approx 1.5\times
10^{-6}$ photons. In contrast, the proposed optimality condition requires
that, on average, one background photon should be incident on the SPAD, per laser cycle. This translates to $\approx 1.5\times 10^{-4}$
photons per bin, which is orders of magnitude higher than extreme attenuation,
and, results in considerably larger signal and SNR.} make the BRC approximately
uniform across all histogram bins, but very low on average, resulting in
extremely low signal. With optimal attenuation, the curve displays some degree
of pile-up, albeit much lower distortion than the case of no attenuation, but
considerably higher values, on average, compared to extreme
attenuation.
Fig.~\ref{fig:bin_receptivity} (c) shows that the optimal attenuation factor is given by the unique maxima location of the minimum value of BRC.
% Maximizing the minimum value of BRC strikes the
% optimal balance between the two goals of making the BRC higher, and, less skewed.
\smallskip

\noindent {\bf Choice of optimality criterion:} Ideally, we should minimize the
root-mean-squared depth error (RMSE or $L^2$) in the design of optimal
attenuation. However, this leads to an intractable optimization problem.
Instead, we choose an upper bound on mean probability of depth error ($L^0)$ as
a surrogate metric, which leads to a closed form minimizer. 
% \textcolor{red}{Reword to say surrogate, and start with $L^2$}.
% $\Upsilon^{\text{opt}}$ in Result is defined as the minimizer of an upper
% bound on $L^0$ depth error (mean probability of depth error). While this
% choice of error metric allows us to derive a closed form minimizer, there is
% no guarantee that other metrics such as $L^2$ error are minimized. Minimizing
% $L^2$ error might be more desirable, since unlike $L^0$, it also penalizes
% the magnitude of errors.
Our simulations and experimental results show that even though
$\Upsilon^{\text{opt}}$ is derived using a surrogate metric, it also
approximately minimizes $L^2$ error, and provides nearly an order of magnitude
improvement in $L^2$ error. \smallskip

\noindent {\bf Estimating $\Phi_\text{bkg}$:} In practice,
$\Phi_\text{bkg}$ is unknown and may vary for each scene point due to distance
and albedo. We propose a simple adaptive algorithm (see supplement) that first
estimates $\lam$ by capturing data over a few initial cycles with the laser
source turned off, and then adapts the attenuation at each point by using the
estimated $\lam$ in Eq.~(\ref{eq:optimality_condition_First}) on a per-pixel
basis. \smallskip

\noindent {\bf Implications of the optimality criterion:}
% \textcolor{red}{Have footnote saying what if $\Upsilon$ depended on signal}.
Note that $\Upsilon^\text{opt}$ is quasi-invariant to scene depths, number of
cycles, as well as the signal strength $\Phi_\text{sig}$ (assuming
$\Phi_\text{sig} \ll B \Phi_\text{bkg}$). Depth-invariance is by design---the
optimization objective in Result~\ref{res:} assumes a uniform prior on the true
depth. As seen from Eq.~(\ref{eq:optimality_condition_First}), this results in an $\Upsilon^\text{opt}$ that doesn't depend on any
prior knowledge of scene depths, and can be easily computed using quantities that are either
known ($\Delta$ and $d_\text{max}$) or can be easily estimated in real-time
($\lam$). The optimal attenuation fraction can be achieved in practice using a
variety of methods including aperture stops, varying the SPAD quantum
efficiency, or with ND-filters.

%Therefore, given a fixed bin-size $\Delta$, $\Upsilon^\text{opt}$ depends only
%on the ambient strength $\Phi_\text{bkg}$, and the depth range $d_\text{max}$.
%In practice, $d_{\text{max}}$ is determined by the application and therefore
%known, and $\lam$ can be estimated (eg. by turning the source off). This has an
%important practical implication: $\Upsilon^\text{opt}$ can be easily computed
%and optimal depth estimation performance can be achieved by attenuating the
%incident flux accordingly (e.g., by varying the aperture or the quantum
%sensitivity of the SPAD, using an ND-filter, etc.), \emph{without} any prior
%knowledge of scene depths~\footnote{Note that this would not be true if
%  $\Upsilon^{\text{opt}}$ had depended on $\muu$, since estimating $\muu$ is a
%non-trivial problem in ambient light-dominated scenes.}. This makes our method
%applicable to many practical situations where depths are not known a priori,
%such as long range LiDAR. Note that although $\Upsilon^{\text{opt}}$ is
%independent of $\Phi_\text{sig}$ and $N$, the depth error still depends on
%both.
\smallskip

\section{Empirical Validation using Simulations}

\begin{figure}
  \centering \includegraphics[width=1.0\columnwidth]{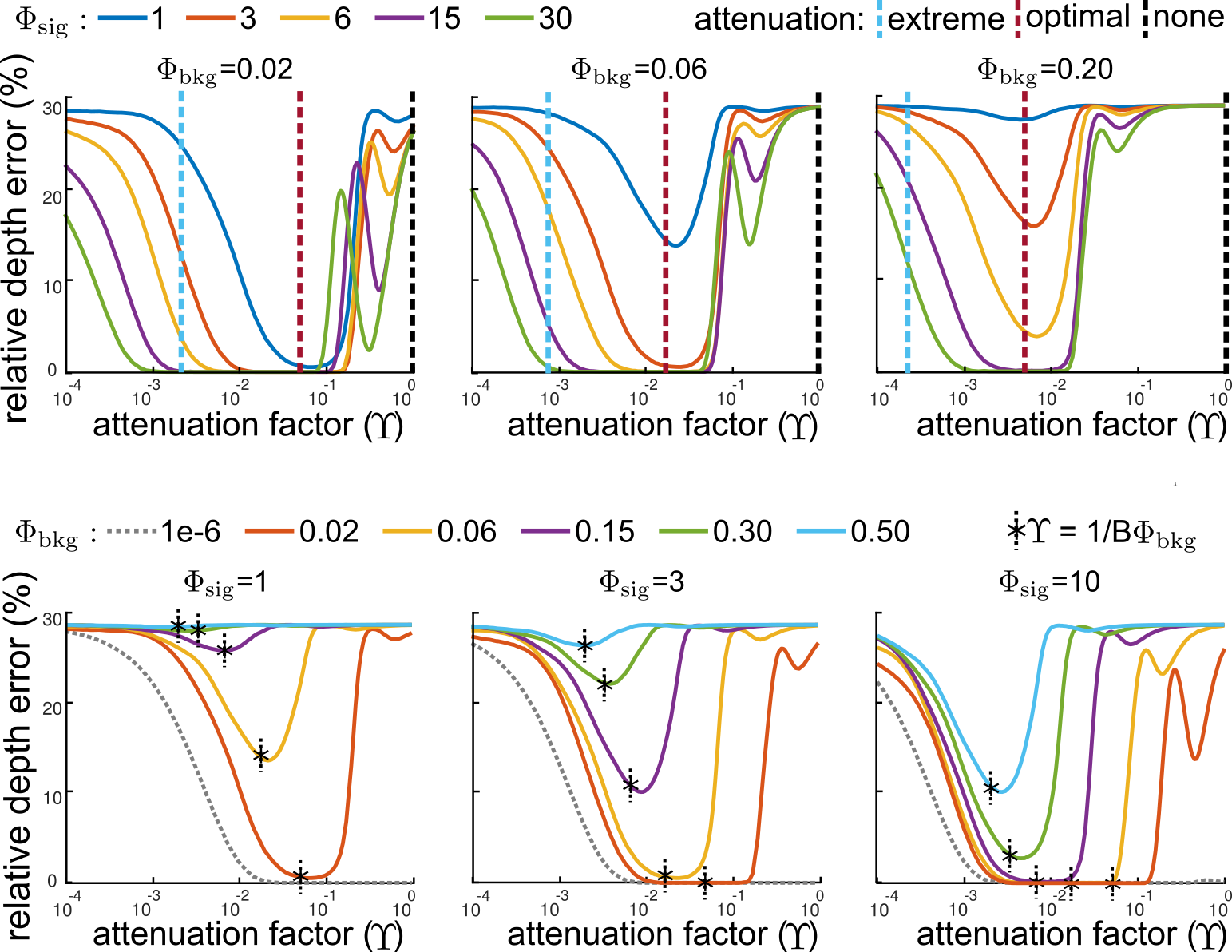}
  \caption{ {\bf Simulation based validation.} (Top row) The values of no,
  extreme, and optimal attenuation are indicated by dotted vertical lines. In
  each of the three plots, the value of optimal attenuation is approximately
  invariant to source power level. The optimal attenuation factor depends only
  on the fixed ambient light level. (Bottom row) For fixed values of source
  power, the optimal attenuation factor increases as ambient light decreases.
  The locations of theoretically predicted optimal attenuation (dotted vertical
  lines) line up with the valleys of the depth error curves.
  \label{fig:single_pixel_simulation_Ucurves}}
  \vspace{-0.1in}
\end{figure}

\begin{figure}[!ht]
\centering \includegraphics[width=1.0\columnwidth]{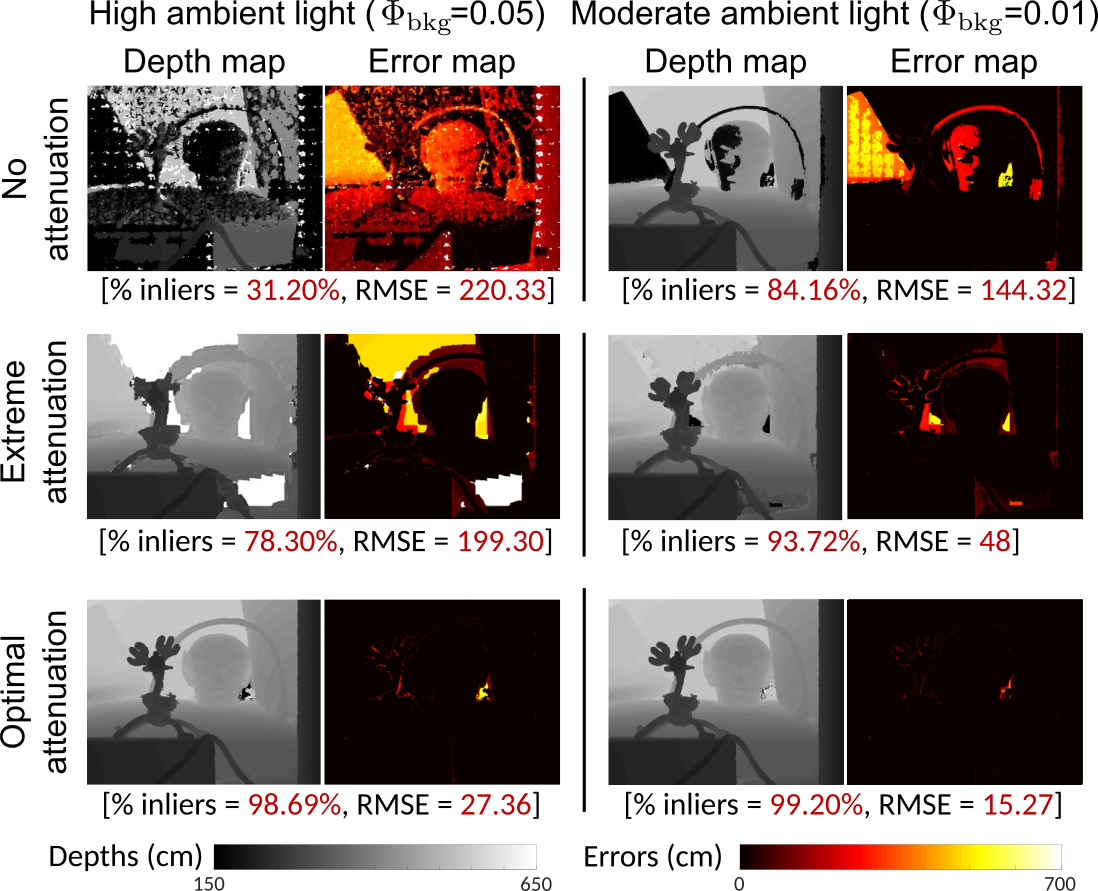}
\caption{ {\bf Neural network based reconstruction for simulations.} Depth and
error maps for neural networks-based depth estimation, under different levels
of ambient light and attenuation. Extreme attenuation denotes average $\Upsilon
B \Phi_\text{bkg} = 0.05$. Optimal attenuation denotes $\Upsilon B
\Phi_\text{bkg} = 1$. \% inliers denotes the percentage of pixels with absolute
error $< \SI{36}{\centi\meter}.$ $\Phi_\text{sig}=2$ for all cases. \label{fig:simulation_dnn}
}
\vspace{-0.1in}
\end{figure}

%\begin{algorithm}
%\caption{Adaptive ND-filtering\label{alg:ndfilter}}
%\begin{myenum}
%  \item Focus the laser source and SPAD detector at a given scene point.
%  \item With the laser power set to zero, acquire a histogram of photon counts
%    $(N'_1, N'_2, \ldots, N'_{B+1})$ over $N'$ laser cycles.
%  \item Estimate the background flux level using:
%    \[
%      \hat \Phi_\text{bkg} = \ln \left( 
%      \frac{\sum_{i=1}^B iN'_i + B N'_{B+1}}{\sum_{i=1}^{B+1}i N_i'-N'}
%      \right).
%    \]
%  \item Set the ND-filtering fraction to $\nicefrac{1}{B\hat\Phi_\text{bkg}}$.
%  \item Set the laser power to the maximum available level and acquire a
%    histogram of photon counts $(N_1, N_2, \ldots, N_{B+1})$ over $N$ laser
%    cycles.
%  \item Estimate the photon flux waveform using
%    Equation~(\ref{eq:coates_corrected_rate_estimate}) and scene depth using
%    Equations~(\ref{eq:time_of_arrival_estimator}) and
%    (\ref{eq:depth_estimator}).
%  \item Repeat for all scene points.
%\end{myenum}
%\end{algorithm}

%\subsection{Validation of the Optimality Criterion}
\paragraph*{Simulated single-pixel mean depth errors:}
We performed Monte Carlo simulations to demonstrate the effect of varying
attenuation on the mean depth error. We
assumed a uniform depth distribution over a range of $1000$ time bins.
%   (corresponds to $\SI{15}{\metre}$ for a bin resolution of $\Delta =
%   \SI{100}{\pico\second}$).
% We considered discrete delta waveforms (Eq.~(\ref{eq:poisson_rates})).
%   $N = 500$ cycles were used which corresponds to an
%   acquisition time of $\SI{100}{\micro\second}$ per measurement, assuming a
%   dead time of $\SI{100}{\nano\second}$ between cycles.
Eq.~(\ref{eq:coates_corrected_rate_estimate}) was used to estimate depths.
% The relative root-mean-squared errors (RMSE) with
% respect to the true depth were averaged over 200 Monte Carlo runs.
Fig.~\ref{fig:single_pixel_simulation_Ucurves} shows plots of the relative
RMSE as a function of attenuation factor $\Upsilon$, for a wide range of $\Phi_{\text{bkg}}$ and $\Phi_{\text{sig}}$ values. 

\begin{figure}
\centering \includegraphics[width=0.9\columnwidth]{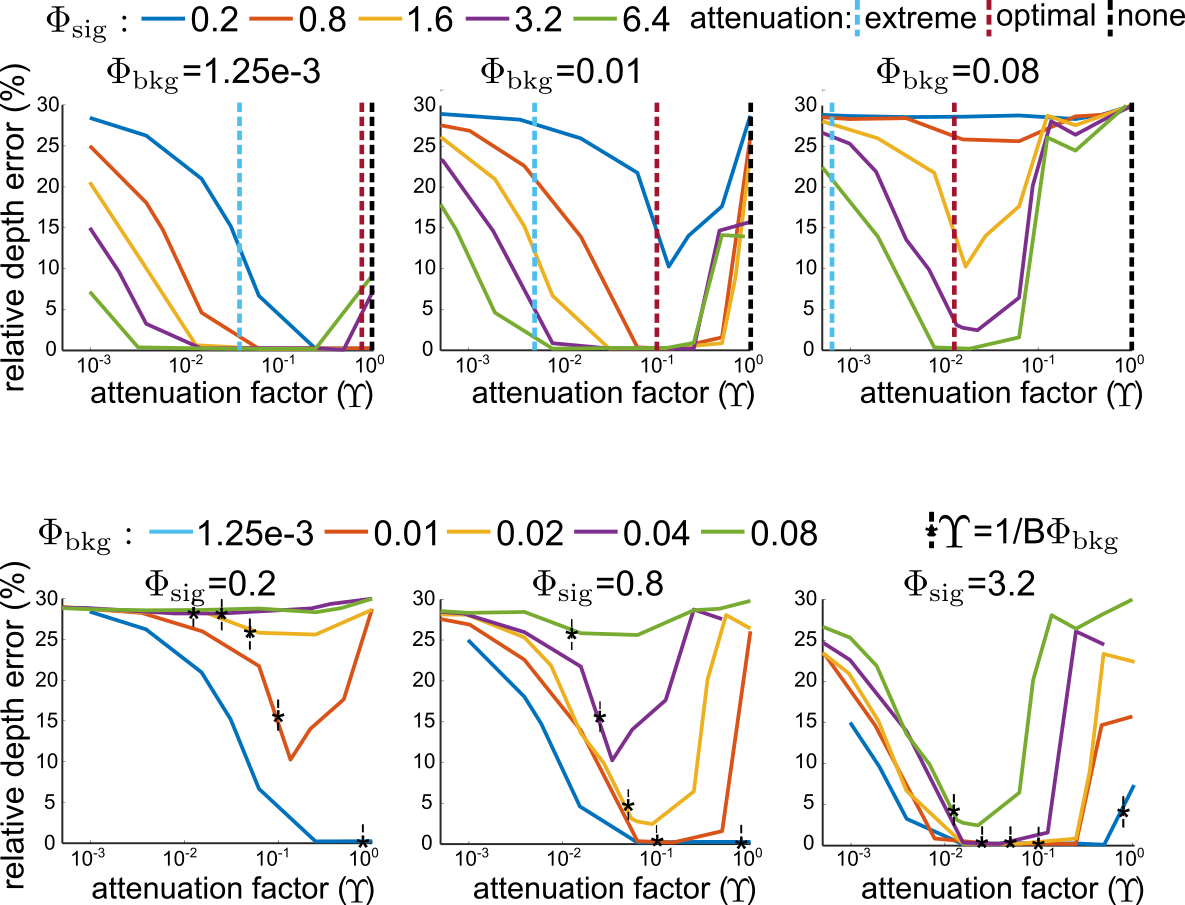}
\caption{ 
{\bf Validation of optimal attenuation using hardware experiments.} These plots
have the same layout as the simulations of
Fig.~\ref{fig:single_pixel_simulation_Ucurves}. As in simulations, the
theoretically predicted locations of the optimal attenuation match the valleys
of the depth error curves.
\label{fig:single_pixel_experiment_Ucurves}}
\vspace{-0.2in}
\end{figure}

Each plot in the top row corresponds to a fixed ambient flux $\Phi_\text{bkg}$.
Different lines in a plot correspond to different signal flux levels
$\Phi_\text{sig}$. There are two main observations to be made here. First, the
optimal attenuation predicted by Eq.~(\ref{eq:optimality_condition})
(dotted vertical line) agrees with the locations of the minimum depth
error valleys
in these error plots.\footnote{As explained in the
supplement, the secondary dips in these error plots at high flux levels are an
artifact of using the Coates's estimator, and are removed by using more
sophisticated estimators such as MAP.}
% \textcolor{red}{This is despite using different metrics for the derivation (upper bound on $L^0$ error) and evaluation (RMSE, i.e., $L^2$ error).}
Second, the optimal attenuation is
quasi-independent of the signal flux $\Phi_\text{sig}$, as predicted by
Eq.~(\ref{eq:optimality_condition}). Each plot in the second row corresponds to
a fixed source flux $\Phi_\text{sig}$; different lines represent different
ambient flux levels. The predicted optimal attenuation align well with the
valleys of respective lines, and as expected, are different for different lines.
% (representing different ambient flux levels).
\smallskip
%\textcolor{red}{We note that even though the optimal flux criterion was
%derived for minimum probability of error, it holds quite well for RMSE which
%is the real metric of interest.} \smallskip

\noindent {\bf Improvements in depth estimation performance:} As seen from all
the plots, the proposed optimal attenuation criterion can achieve up to \emph{1
order of magnitude} improvement in depth estimation error as compared to
extreme or no attenuation. Since most valleys are relatively flat, in general,
the proposed approach is robust to uncertainties in the estimated background
flux, and thus, can achieve high depth precision across a wide range of
illumination conditions. \smallskip

\noindent{\bf Validation on neural networks-based depth estimation:} Although
the optimality condition is derived using an analytic pixel-wise depth
estimator~\cite{Coates}, in practice, it is valid for state-of-the-art deep
neural network (DNN) based methods that exploit spatio-temporal correlations in
natural scenes. We trained a convolutional DNN \cite{Lindell} using simulated
pile-up corrupted histograms, generated using ground truth depth maps from
the NYU depth dataset V2 \cite{Silberman:ECCV12}, and tested
on the Middlebury dataset \cite{scharstein2007learning}. For each combination of ambient flux, source flux
and attenuation factor, a separate instance of the DNN was trained on
corresponding training data, and tested on corresponding test data to
ensure a fair comparison across the different attenuation methods.

Fig.~\ref{fig:simulation_dnn} shows depth map reconstructions at different
levels of ambient light. If no attenuation is used with high ambient light, the
acquired data is severely distorted by pile-up, resulting in large depth
errors. With extreme attenuation, the DNN is able to smooth out the effects of shot noise, but results in blocky edges.
% Objects
% that are farther away or have low albedo suffer from large errors. 
With
optimal attenuation, the DNN successfully recovers the depth map with
considerably higher accuracy, at all ambient light levels.

%The supplementary report shows additional results with state-of-the-art depth
%estimation techniques that incorporate prior knowledge about the laser
%waveform shape in a MAP framework and methods that use spatial regularizers to
%remove ambient photons \cite{Rapp}.

\begin{figure}[!t]
\centering \includegraphics[width=0.9\columnwidth]{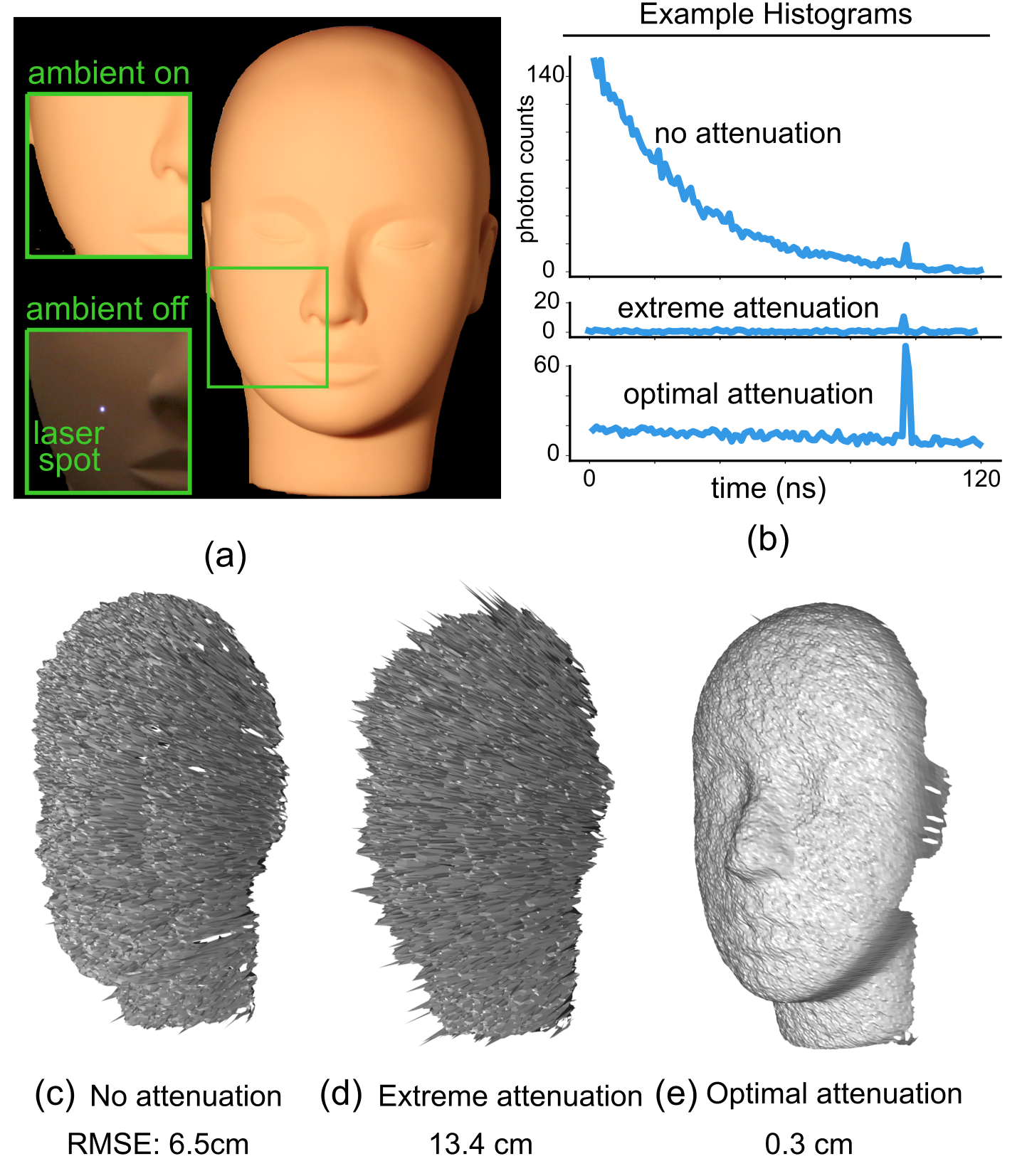}
\caption{ {\bf 3D reconstruction of a mannequin face}
(a) A mannequin face illuminated by bright ambient light. The laser spot is barely visible. (b)
Representative histograms acquired from the laser position shown in (a). 
% Very
% low photon counts are obtained when using extreme attenuation and the peak
% corresponding to the scene depth is barely visible. With optimal filtering,
% there is some pile up, but the peak is visible. With no filtering we observe
% extreme pile up and the peak gets buried in the exponentially decaying tail of
% the histogram.
With extreme and no attenuation, the peak corresponding to the scene depth is
barely identifiable. With optimal attenuation, the peak can be extracted
reliably. (c-d) The depth reconstructions using no and extreme attenuation
suffer from strong pile-up and shot noise, (e) Optimal attenuation achieves an
order of magnitude higher depth precision, even enabling recovery of fine details.
\label{fig:mannequin_face}}
\vspace{-0.2in}
\end{figure}

\begin{figure*}[!ht]
\centering \includegraphics[width=0.93\textwidth]{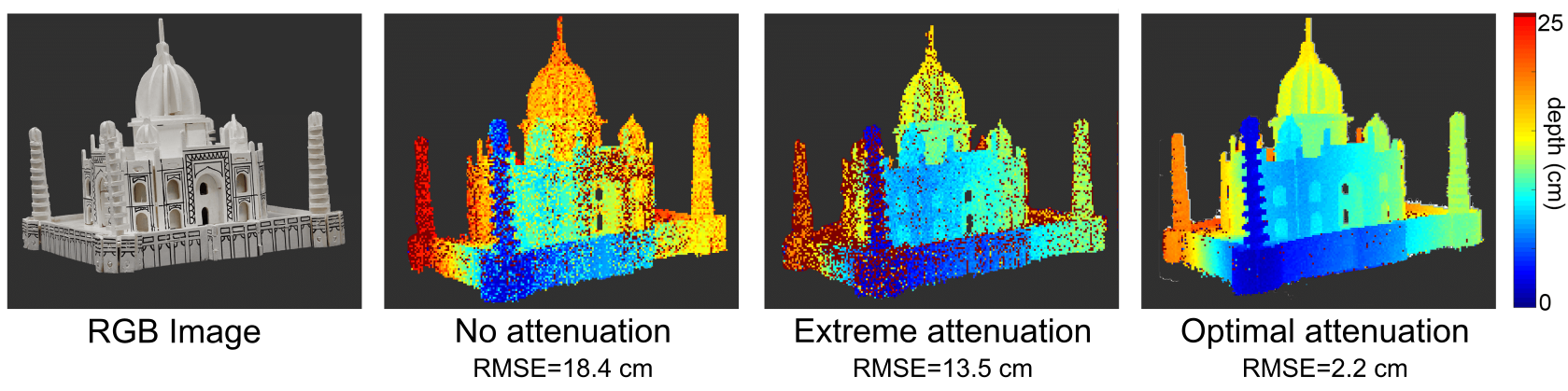}
\caption{ {\bf Depth estimation with varying attenuation.} The
average ambient illuminance of the scene was $\SI{15000}\lux$. With no
attenuation, most parts are affected by strong pile-up, resulting in several
outliers. For extreme attenuation, large parts of the scene have very low SNR.
In contrast, optimal attenuation achieves high depth estimation performance for
nearly the entire object. (\SI{15}{\meter} depth offset removed.) \label{fig:tajmahal}}
\vspace{-0.1in}
\end{figure*}

\begin{figure*}[!ht]
\centering \includegraphics[width=0.92\textwidth]{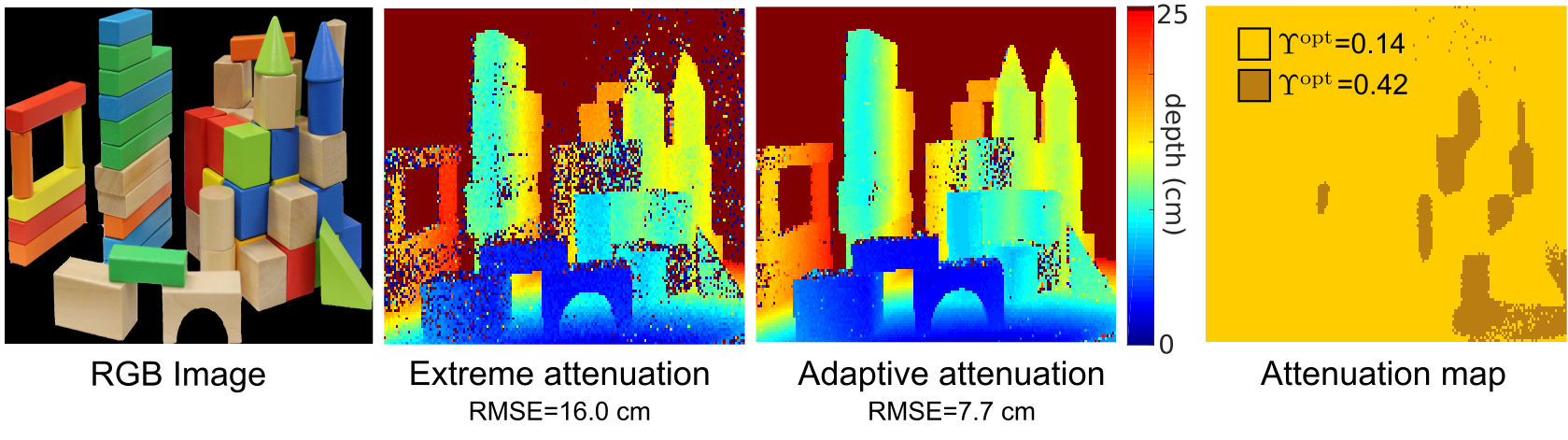}
\caption{ {\bf Ambient-adaptive $\Upsilon^\text{opt}$.} This scene has large
ambient brightness variations, with both brightly lit regions (right) and
shadows (left). Pixel-wise ambient flux estimates were used to adapt
the optimal attenuation, as shown in the attenuation map.
The resulting reconstruction achieves accurate estimates, both in shadows and
brightly lit regions. (\SI{15}{\meter} depth offset removed.) \label{fig:blocks}}
\vspace{-0.2in}
\end{figure*}

\section{Hardware Prototype and Experiments\label{sec:expt}}
Our hardware prototype is similar to the schematic shown in
Fig.~\ref{fig:pulsed_lidar}. We used a $\SI{405}{\nano\meter}$ wavelength,
pulsed, picosecond laser (PicoQuant LDH P-C-405B) and a co-located fast-gated
single-pixel SPAD detector \cite{buttafava2014spad} with a
$\SI{200}{\nano\second}$ dead time. The laser repetition rate was set to 5~MHz
corresponding to $d_\text{max}=\SI{30}{\meter}$. Photon timestamps
were acquired using a TCSPC module (PicoQuant HydraHarp 400). Due to practical
space constraints, various depths covering the full $\SI{30}{\meter}$ of
unambiguous depth range in Fig.~\ref{fig:single_pixel_experiment_Ucurves} were
emulated using a programmable delayer module (Micro Photon Devices PSD).
Similarly, all scenes in Figs.~\ref{fig:mannequin_face},
\ref{fig:tajmahal} and \ref{fig:blocks} were provided with a depth offset of
$\SI{15}{\meter}$ using the PSD, to mimic long range LiDAR.
\smallskip

\noindent{\bf Single-pixel Depth Reconstruction Errors:}
Fig.~\ref{fig:single_pixel_experiment_Ucurves} shows the relative depth errors
that were experimentally acquired over a wide range of ambient and source flux
levels and different attenuation factors. These experimental curves follow the
same trends observed in the simulated plots of
Fig.~\ref{fig:single_pixel_simulation_Ucurves} and provide experimental
validation for the optimal flux criterion in the presence of non-idealities
like jitter and afterpulsing effects, and for a non-delta waveform.
\smallskip

\noindent{\bf 3D Reconstructions with Point Scanning:}
Figs.~\ref{fig:mannequin_face} and~\ref{fig:tajmahal} show 3D reconstruction
results of objects under varying attenuation levels, acquired by
raster-scanning the laser spot with a two-axis galvo-mirror system (Thorlabs
GVS-012).
% Extreme filtering is corrupted by shot noise and will require
% much longer acquisitions to achieve acceptable depth accuracy. The histograms
% are severely corrupted by pileup and in the absence of any filtering the depth
% map appears extremely noisy.
It can be seen from the histograms in Fig.~\ref{fig:mannequin_face} (b) that
extreme attenuation almost completely removes pile-up, but also reduces the
signal to very low levels. In contrast, optimal attenuation has some residual
pile-up, and yet, achieves approximately an order of magnitude higher depth
precision as compared to extreme and no attenuation. Due to relatively uniform
albedos and illumination, a single attenuation factor for the whole scene was
sufficient.
% Optimal filtering successfully recovers depth
% details such as eyelids and lips.
% \textcolor{red}{The depth maps were
% estimated using point-wise Coates's estimation, and the data was acquired for
% $4000$ cycles for all attenuation levels.}
%\textcolor{red}{To get a reasonable SBR in the confocal setup, we used relatively low ambient flux levels. However, even in this benign setting, there was significant pile-up and using optimal attenuation led to dramatic improvements.}  
% \textcolor{red}{Figure~\ref{fig:tajmahal} shows the effect of different
% attenuation levels on reconstructed depth maps under strong ambient light. With
% no attenuation as well as with linear regime attenuation (5\% photon flux),
% many pixels have noisy depth estimates, for reasons discussed earlier. With
% optimal attenuation, there is almost no noise in the reconstruction, and even
% fine details are recovered well. For this experiment, $2000$ cycles were
% captured.}

Fig.~\ref{fig:blocks} shows depth maps for a complex scene containing a wider
range of illumination levels, albedo variations and multiple objects over a
wider depth range. The optimal scheme for the ``Blocks'' scene adaptively
chooses different attenuation factors for the parts of the scene in direct and
indirect ambient light.\footnote{ In this proof-of-concept, we acquired
multiple scans at different attenuations, and stitched together the final
depth map in post-processing.  } Adaptive attenuation enables depth
reconstruction over a wide range of ambient flux levels.
% The ``Tabletop'' scene consists of multiple objects and shows that our
% method can accurately reconstruct depth maps for scenes with a wide range of
% albedo and depth variations.
% Edited for Arxiv
% {\bf See supplementary material for more results.}

\section{Limitations and Future Outlook}
\noindent {\bf Achieving uniform depth precision across depths:} The optimal
attenuation derived in this paper results in a high and relatively less skewed
BRC (as shown in Fig.~\ref{fig:bin_receptivity}), resulting in high depth
precision across the entire depth range. However, since the optimal curve has
some degree of pile-up and is monotonically decreasing, later bins
corresponding to larger depths still incur larger errors. It may be possible to
design a time-varying attenuation scheme that gives uniform depth estimation
performance. \smallskip

\noindent {\bf Handling non-impulse waveforms:} Our analysis assumes ideal
delta waveform, as well as low source power, which allows ignoring the effect
of pile-up due to the source itself. For applications where source power is
comparable to ambient flux, a next step is to optimize over non-delta
waveforms~\cite{Heide:2018:subpicosecond} and derive the optimal flux
accordingly. \smallskip

%A next step is to extend the presented analysis to general waveform estimation for applications where indirect light components are significant, such as NLOS imaging~\cite{OToole2018}.  

\noindent {\bf Multi-photon SPAD LiDAR:} With recent improvements in detector
technology, SPADs with lower dead times (tens of ns) can be
realized, which enable capturing more than one photon per laser cycle.
This includes multi-stop TCSPC electronics and SPADs that can
be operated in the free-running mode, for which imaging models and estimators
have been proposed recently \cite{rapp2018dead, isbaner2016dead}. An interesting future
direction is to derive optimal flux criterion for such multi-photon SPAD-based
LiDARs.

{\small
\bibliographystyle{ieee_fullname}
\bibliography{egbib}
}

\clearpage
\onecolumn
\renewcommand{\figurename}{Supplementary Figure}
\renewcommand{\thesection}{S. \arabic{section}}
\renewcommand{\theequation}{S\arabic{equation}}
\setcounter{figure}{0}
\setcounter{section}{0}
\setcounter{equation}{0}
\setcounter{page}{1}

\begin{center}
%\begin{tabular}{c}
\huge Supplementary Document for\\[0.2cm]
\huge ``\mytitle'' \\[2.1cm]
\normalsize Anant Gupta, Atul Ingle, Andreas Velten, Mohit Gupta.\\[0.5cm]
Correspondence to: agupta225@wisc.edu
% \normalsize Anonymous CVPR Submission\\
% Paper ID 1624
%\end{tabular}
\end{center}

% \subsection{Training CNN details}

\section{Computational Pile-up Correction via Analytic Inversion (Coates's Method)}
Theoretically, it is possible to ``undo'' the pile-up distortion in the measured histogram by analytically inverting the SPAD image formation model. This method, also called as Coates's correction in the paper~\cite{Coates}, provides a closed form expression for the true incident waveform $r_i$ as a function of the measured (distorted) histogram $N_i$ (Section 4.1 of the main paper). 

In this section, we provide theoretical justification for using this method, and show that it is equivalent to computing the maximum likelihood estimate (MLE) of the true incident waveform, and therefore, under certain settings, provably optimal. This result was also proved in \cite{pediredla2018signal}, and is provided here for completeness. This method has an additional desirable property of providing unbiased estimates of the incident waveform. Furthermore, this method assumes no prior
knowledge about the shape of the incident waveform, and thus, can be used to estimate arbitrary incident waveforms, including those with a single dominant peak (e.g., typically received by a LIDAR sensor) for estimating scene depths. 

%\noindent{\bf Summary:} We provide theoretical justification for using the Coates's estimator, and show that it is optimal for estimating arbitrary waveforms. Furthermore, since estimating the waveform is a precursor to estimating the location of the peak (i.e. the time-of-flight), this leads to a reliable depth estimate.

%In Coates's correction~\cite{Coates}, the analytical formula for the probability  estimate $\widehat q_i$ (where $q_i = 1 - e^{-r_i}$) is obtained by simply  inverting the expressions for the photon counts $N_i$ in terms of the true  probabilities $q_i$. Here we show that the simple inversion is equivalent to  finding the maximum likelihood estimate (MLE) of the probability vector $(q_1, q_2, ..., q_B) \in [0, 1]^B$. The Coates's method assumes no prior knowledge about the shape of the waveform to be estimated, it can have an arbitrary shape.  By the asymptotic efficiency of MLE,\footnote{More explicitly, the MLE achieves a lower bound on the variance of all unbiased estimators, known as the Cramer-Rao bound.} this is the best possible estimator for the probabilities $q_i$ in this setting. 

\subsection{Derivation of MLE}
In any given laser cycle, the detection of a photon in the $i^\text{th}$ bin is
a Bernoulli trial with probability $q_i = 1 - e^{-r_i}$, conditioned on no
photon being detected in the preceding bins. Therefore, in $N$ cycles, the
number of photons $N_i$ detected in the $i^\text{th}$ bin is a binomial
random variable when conditioned on the number of cycles with no photons
detected in the preceding bins.
\begin{equation}
\label{supp1}
N_i \;\vert\; D_i \sim \mathsf{Binomial}\left(D_i, q_i\right),
\end{equation}
where $D_i$ is the number of cycles with no photons detected in bins $1$ to
$i-1$ and can be expressed in terms of the histogram counts as:
\[ D_i = N - \sum_{j=1}^{i-1} N_j. \]
Therefore, the likelihood function of the probabilities $(q_1, q_2, ..., q_B)$ is given by:
\begin{align*}
 L(q_1, q_2, ..., q_B) &= \P(N_1, N_2, ..., N_B | q_1, q_2, ..., q_B) \\
 &= \P(N_1 | q_1) \prod_{i=2}^B \P(N_i | q_i, N_1, N_2, ..., N_{i-1}) \\
 &= \P(N_1 | q_1, D_1) \prod_{i=2}^B \P(N_i | q_i, D_i). 
\end{align*}
by the chain rule of probability, and using the fact that $N_i$ only depends on
its probability $q_i$ and preceding histogram counts. Each term of the product
is given by the binomial probability from Eq.~(\ref{supp1}). Since each $q_i$
only affects a single term, we can calculate its MLE separately as:
\begin{align}
 \widehat q_i &= \argmax_{q_i}\,\, \P(N_i | q_i, D_i) \nonumber \\
 &= \argmax_{q_i} \,\, {D_i \choose N_i} q_i^{N_i} (1 - q_i)^{D_i - N_i} \nonumber \\
 &= \frac{N_i}{D_i}
 = \frac{N_i}{N - \sum_{j=1}^{i-1} N_j}. \label{justcalc}
\end{align}
% \[ = p_b \pm \sqrt{\frac{p_b(1-p_b)}{N - \sum_{i=1}^{b-1} N_i}} \] where the
% error in the estimate of $p_b$ is computed using the standard deviation of
% the conditional binomial random variable $N_b \;\vert\; N_1..N_{b-1}$,
% ignoring the uncertainty of the number of cycles $N - \sum_{i=1}^{b-1} N_i$
% itself.
\subsection{Calculating the bias of Coates's corrected estimates}
From Eq.~(\ref{justcalc}) for the MLE, we have for each $1 \leq i \leq B$:
\[ \E[\widehat q_i] = \E{\left[\frac{N_i}{D_i}\right]} \]
By the law of iterated expectations:
\begin{align}
  \E[ q_i ] &=  \E \left[ \E{\left[\left.\frac{N_i}{D_i}\right| N_1, N_2, ..., N_{i-1}\right]} \right]\\
            &= \E \left[\frac{q_i D_i}{D_i} \right] = q_i
\end{align}
where the last step uses the mean of the binomial distribution.

Therefore, $\widehat q_i$ is an unbiased estimate of $q_i$. By combining the
expression for $\widehat q_i$ with $\widehat{r}_i = \ln{\left(\frac{1}{1 -
  \widehat{q}_i}\right)}$, we get the Coates's formula mentioned in Section 4.1
  of the main text.
%\newline \newline

\section{Derivation of the Optimal Attenuation Factor $\Upsilon^\text{opt}$}
In this section, we derive the expression for optimal attenuation factor
$\Upsilon^\text{opt}$ in terms of the bin receptivities $C_i$. We first compute
some properties of the Coates's estimator which are needed for the derivation.
Then we derive an upper bound on the probability that Coates's estimator
produces the incorrect depth. This upper bound is a function of $\Upsilon$.
The optimal $\Upsilon$ then follows by minimizing the upper bound.

We assume that the incident waveform is the sum of a constant ambient light
level $\Phi_{\text{bkg}}$ and a single laser source pulse of height
$\Phi_{\text{sig}}$. Following the notation used in the main text, we have:
\[ 
  r_i = \Phi_{\text{sig}} \delta_{i, r} + \Phi_{\text{bkg}}.
\]
Furthermore, we assume that $r_i$ is small enough so that $q_i =  1 - e^{-r_i}
\approx r_i$.\footnote{Note that this assumption is different from low flux
assumption used in the linear operation regime which requires even lower
flux levels satisfying $r_i \ll 1/B$.}

\subsection{Variance of Coates's estimates}
From the previous section, the Coates's estimator is given by:
\[ \widehat q_i = \frac{N_i}{D_i} \]
and the Coates's time-of-flight estimator is given by:
\[ \widehat \tau = \argmax_{i} \widehat q_i \]
Note that locating the peak in the waveform is equivalent to locating the
maximum $q_i$.  From the previous section, we know that $\E [\widehat q_i] =
q_i $.  Intuitively, this means that the estimates of $q_i$ are correct on
average, and we can pick the maximum $\widehat q_i$ to get the correct depth, on
average.  However, in order to bound the probability of error, we need
information about variance of the estimates. Let $\sigma^2_{i}$ denote the
diagonal terms and $\sigma^2_{i,j}$ denote the off-diagonal terms of the
covariance matrix of $(\widehat{q}_1, \widehat{q}_2, \ldots, \widehat{q}_B)$.
We have:
\begin{align}
\sigma^2_i &= \E[(\widehat q_i - q_i)^2] \nonumber \\
& = \E \left[ {\left( \frac{N_i}{D_i} - q_i \right)}^2 \right] \nonumber \\
& = \E \left[ \E \left[ \left. {\left(\frac{N_i}{D_i} - q_i \right)}^2 \right| D_i\right] \right] \label{step1} \\
& = \E \left[\frac{q_i(1 - q_i)}{D_i} \right] \label{step2}
\end{align}
where Eq.~(\ref{step1}) uses the law of iterated expectations and
Eq.~(\ref{step2}) uses the variance of the binomial distribution. Note that
$D_i$ is also a binomial random variable, therefore,
\begin{equation}
\label{derivation}
\sigma_i^2 = \E\left[\frac{q_i(1 - q_i)}{D_i}\right] \approx \frac{q_i(1 - q_i)}{\E{[D_i]}}
\end{equation}
where in the last step, we have interchanged expectation and reciprocal. This
can be seen to be true when $D_i$ is large enough so that $D_i \approx
D_{i}+1$, by writing out $\E[\nicefrac{1}{D_i+1}]$ explicitly. Recalling the
definition of $D_i$ and using the mean of the multinomial distribution, we
have:
\[
 \E{[D_i]} = \E\left[N - \sum_{j=1}^{i-1} N_j\right] = N \left(1 - \sum_{j=1}^{i-1} p_j \right) = \frac{N p_i}{q_i}
\]
where the last step follows after some algebraic manipulation involving the
definition of $p_i$.  Substituting this into Eq.~(\ref{derivation}) and using
the definition of bin receptivity, we get:
\[
  \sigma^2_i = \frac{q_i^2 (1-q_i)}{N p_i} = \frac{q_i^2 (1-q_i)r}{N C_i r_i} \approx \frac{r_i r}{N C_i}
\]
since $r_i \approx q_i \ll 1$ by assumption.

Next we compute $\sigma_{i,j}, i \neq j$. Without loss of generality, assuming
$i<j$, we have:
\begin{align}
\sigma^2_{i,j} &= \E\left[(\widehat q_i - q_i)(\widehat q_j - q_j)\right] \nonumber \\
 &=  \mathbb{E}_{N_1,N_2,...,N_i}\left[(\widehat q_i - q_i)\E_{N_{i+1},...,N_B|N_1,...N_i}(\widehat q_j - q_j)\right] \nonumber \\
 &= \mathbb{E}_{N_1,N_2,...,N_i}\left[(\widehat q_i - q_i)\E_{N_{j},D_j|N_1,...N_i}(\widehat q_j - q_j)\right] \label{step3} \\
 &=\mathbb{E}_{N_1,N_2,...,N_i}\left[(\widehat q_i - q_i)\E_{D_j|N_1,...N_i}\E_{N_j | D_j}(N_j/D_j - q_j)\right] = 0 \label{step4}
\end{align}
where Eq.~(\ref{step3}) uses the fact that $\widehat q_j = N_j/D_j$ only
depends on $N_j$ and $D_j$, and Eq.~(\ref{step4}) uses the fact that the
innermost expectation is zero. Therefore, $\sigma^2_{i,j} = 0$ and $\widehat
q_i$ and $\widehat q_j$ are uncorrelated for $i \neq j.$

\subsection{Upper bound on depth error probability}
To ensure that the estimated depth is correct, the bin corresponding to the
actual depth should have the highest Coates-corrected count. Therefore, for a
given true depth $\tau$, we want to minimize the probability of error
$\P(\widehat \tau_{\text{Coates}} \neq \tau).$
\begin{align*}
\P(\widehat \tau_{\text{Coates}} \neq \tau) &= \P\left(\bigcup_{i\neq \tau} (\widehat q_i > \widehat q_{\tau})\right)\\
&\leq \sum_{i \neq \tau} \P\left(\widehat q_i > \widehat q_{\tau}\right) \\
           &= \sum_{i \neq \tau} \P\left(\widehat q_i - \widehat q_{\tau} > 0 \right).
\end{align*}
Note that $\widehat q_i - \widehat q_\tau$ has a mean $q_i - q_\tau$ and variance $\sigma^2_i +
\sigma^2_\tau$, since they are uncorrelated. For large $N$, by the central
limit theorem, we have:
\[ \widehat q_i - \widehat q_\tau \sim \mathcal{N}(q_i - q_\tau, \sigma^2_i + \sigma^2_\tau). \]
%\[ = \sum_{i \neq \tau} \P\left(\widehat p(i) > \widehat p(\tau)\right) \]
Using the Chernoff bound for Gaussian random variables, we get:
% Assuming that the estimated normalized rates $\widehat \rho_i$ are $\sigma_{\widehat \rho_i}$-subgaussian with means $\rho_i$ and independent, we get:
\begin{align*}
 \P(\widehat q_i > \widehat q_\tau) &\leq \frac{1}{2}\exp{\left( - \frac{(q_i - q_\tau)^2}{2 (\sigma^2_i + \sigma^2_\tau)} \right)} \\
 &\approx \frac{1}{2} \exp{\left( - \frac{N (r_i - r_\tau)^2}{2 (\frac{r_i r}{C_i} + \frac{r_\tau r}{C_\tau})} \right)} \\ 
 &= \frac{1}{2} \exp{\left( - \frac{N (\frac{r_i}{r} - \frac{r_\tau}{r})^2}{2 (\frac{r_i}{r C_i} + \frac{r_\tau}{r C_\tau})} \right)} \\
 &= \frac{1}{2}\exp{\left( - \frac{N{\left(\frac{\Phi_\text{sig}}{B \Phi_\text{bkg} + \Phi_\text{sig}}\right)}^2}{2 \left(\frac{\Phi_\text{bkg}}{(B \Phi_\text{bkg} + \Phi_\text{sig})C_i} + \frac{\Phi_\text{bkg}+\Phi_\text{sig}}{(B \Phi_\text{bkg} + \Phi_\text{sig})C_\tau}\right)} \right)} 
 \end{align*}
where the last step uses the definition of $r_i$. Since we are interested in
the case of high ambient light and low source power, we assume $\Phi_\text{sig}
\ll B\Phi_\text{bkg}$. The above expression then simplifies to:
\[ \P(\widehat q_i > \widehat q_\tau) \leq \frac{1}{2}\exp{\left( - \frac{\frac{N}{B}{\theta^2}}{2 \left(\frac{1}{C_i} + \frac{1 + \theta}{C_j}\right)} \right)} \]
where $\theta = \Phi_\text{sig} / \Phi_\text{bkg}$ denotes the SBR. Assuming a
uniform prior on $\tau$ over the whole depth range, we get the following upper bound
on the average probability of error:
\begin{align*}
 \frac{1}{B} \sum_{\tau=1}^B \P(\widehat \tau_{\text{Coates}} \neq \tau)  &\leq \frac{1}{B}\sum_{\tau=1}^B \sum_{i \neq \tau} \frac{1}{2}\exp{\left( - \frac{\frac{N}{B}\theta^2}{2 \left(\frac{1}{C_i} + \frac{1 + \theta}{C_\tau}\right)} \right)} \\
 &\approx \frac{1}{B}\sum_{\tau=1}^B \sum_{i=1}^B \frac{1}{2}\exp{\left( - \frac{\frac{N}{B}\theta^2}{2 \left(\frac{1}{C_i} + \frac{1 + \theta}{C_\tau}\right)} \right)}
\end{align*}
We can minimize the probability of error indirectly by minimizing this upper
bound. The upper bound involves exponential quantities which will be dominated
by the least negative exponent. Therefore, the optimal attenuation is given by:
\begin{align*}
 \Upsilon^{\text{opt}} &= \argmin_{\Upsilon} \frac{1}{B}\sum_{i,\tau=1}^B \frac{1}{2}\exp{\left( - \frac{{\frac{N}{B} \theta^2}}{2 \left(\frac{1}{C_i} + \frac{1 + \theta}{C_\tau}\right)} \right)} \\
 &\approx \argmin_{\Upsilon} \max_{i, \tau} \left[ \frac{1}{2}\exp{\left( - \frac{{\frac{N}{B} \theta^2}}{2 \left(\frac{1}{C_i} + \frac{1 + \theta}{C_\tau}\right)} \right)} \right] \\
 &= \argmax_{\Upsilon} \min_{i} C_i
 \end{align*}
The last step is true since the term inside the exponent is maximized for $i =
\tau = \argmin_i C_i(r)$. Furthermore, the expression depends inversely on
$C_i$ and $C_\tau$, and all other quantities ($N, B, \theta$) are independent
of $\Upsilon$. Therefore, minimizing the expression is equivalent
to maximizing the minimum bin receptivity.

\subsection{Interpretation of the optimality criterion as a geometric tradeoff}
We now provide a justification of our intuition that the optimal flux should
make the BRC both uniform and high on average. The optimization objective $\min_i C_i(\Upsilon)$
(Eq.~(\ref{eq:optimality_condition_First})) of Section 4.2 can be decomposed as:
\begin{align*}
\min_i C_i(\Upsilon) &= C_B(\Upsilon) = B \, (1-e^{-\Upsilon \, \Phi_\text{bkg} }) \, e^{-(B-1) \Upsilon \, \Phi_\text{bkg}} \\
              & \hspace{-0.6in} = \left[1 - e^{-\Upsilon B \Phi_\text{bkg}}\right]
                \frac{1}{\frac{1}{B(1-e^{-\Upsilon \Phi_\text{bkg}}) e^{(-\Upsilon B \Phi_\text{bkg})}} -
                \frac{1}{B(1-e^{-\Upsilon \Phi_\text{bkg}}) }} \\
              &\hspace{-0.6in} =  \underbrace{\frac{1}{B}\sum_{i=1}^B C_i(\Upsilon)}_\text{Mean receptivity} 
              \left(\underbrace{\frac{1}{C_{B}(\Upsilon)} - \frac{1}{C_1(\Upsilon)}}_\text{Skew}\right)^{-1}.
\end{align*}
The first term is the mean receptivity (area under the BRC). The second term is
a measure of the non-uniformity (skew) of the BRC. Since the optimal $\Upsilon$
maximizes the objective $\min_i C_i(\Upsilon)$, which is the ratio of mean
receptivity and skew, it simultaneously achieves low distortion and large mean
values.
\newline \newline
\noindent{\bf Summary:} We derived the optimal flux criterion of Section $5$ in
the main text, using an argument about bounding the mean probability of error.
The expression for optimal attenuation depends on a geometric quantity, the bin
receptivity curve, which also has an intuitive interpretation.

% \subsubsection{Optimality condition}
% The metric defined above gives us a way to compare different bin receptivity curves. As discussed earlier, we are interested in the optimal photon flux that should be incident on a SPAD for the application of LIDAR. Since $C(i)$ is parameterized in terms of the total flux $r$, we find:
% \[ r_{opt} = \argmin_{r} M_{C_{r}} \]
% Substituting the parameterized definition of $C_{r}$ from above, we get:
% \[ r_{opt} = \argmax_{r} \min_i C_{r}(i) \]
% \[ = \argmax_{r} C_{r}(n). \]
% Intuitively, the last bin is affected the most by pile-up, and therefore has the highest uncertainty in its rate estimate. Minimizing the error metric boils down to minimizing that uncertainty, and therefore maximizing $1 / \sigma_{\widehat r_i}$.

\section{Alternative computational methods for pile-up correction}
In this section we present depth estimation methods that can be used as
alternatives to the Coates's estimator in situations where additional
information about the scene is available.
\newline\newline
\noindent\textbf{Suboptimality of Coates's method for restricted waveform types:}
In our analysis of depth estimation in SPADs, we used the Coates's estimator
for convenience and ease of exposition. The Coates's method estimates depth
indirectly by first estimating the flux for each histogram bin. Although this
is optimal for depth estimation with arbitrary waveforms, it is suboptimal in
our setting where we assume some structure on the waveform.
First, it does not utilize the shared parameter space of the incident waveform,
which can be described using just three parameters: background flux
$\Phi_\text{bkg}$, source flux $\Phi_\text{sig}$ and depth $d$. Instead, the
Coates method allows an arbitrary waveform shape described by $B$ independent
parameters for the flux values at each time bin.  Moreover, it does not assume
any prior knowledge of $\Phi_\text{bkg}$ and $\Phi_\text{sig}$.
\newline\newline
\noindent{\bf MAP and Bayes estimators:} 
In the extreme case, if we assume $\Phi_\text{bkg}$ and $\Phi_\text{sig}$ are
known, the only parameter to be estimated is $d$. We can then explicitly
calculate the posterior distribution of the depth using Bayes's rule:
\[ 
\P(d|N_1,N_2,..., N_B) = \frac{\P(d) \P(N_1,N_2,...,N_B|d)}{\P(N_1,N_2,...,N_B)}.
\]
Assuming a uniform prior on depth, this can be simplified further: 
\begin{align}
\P(d|N_1,\ldots, N_B) &= \frac{\P(N_1,\ldots,N_B|d)}{\sum_{i=1}^n\P(N_1,\ldots,N_B|i)} \nonumber \\
&\propto \P(N_1,\ldots,N_B|d) \nonumber \\
&= \prod_{i=1}^B (q_{i|d})^{N_i} (1 - q_{i|d})^{N - \sum_{j=0}^{i-1} N_j} \nonumber \\
& = \exp{\left\{\sum_{i=1}^B N_i \ln{(q_{i|d})} + \sum_{i=1}^B   D_i \ln{(1 - q_{i|d})} \right\}} \label{corr}
\end{align}
where $q_{i|d}$ denotes the incident photon probability at the $i^\text{th}$
bin when the true depth is $d$. Note that $q_{i|d}$ for different depths are
related through a rotation of the indices
$q_{i|d} = q_{(i - d) \bmod B | 0}.$
Therefore, the expression in the exponent of Eq.~(\ref{corr}) can be computed
efficiently by a sum of two correlations.  The Bayes and MAP estimators are
then given by the mean and mode of the posterior distribution respectively.
\newline \newline
\noindent{\bf Advantages of MAP Estimation:} It can be shown that Bayes and MAP
estimators are optimal in terms of mean squared loss and 0-1 loss respectively
\cite{bayesian}. Unlike the Coates method, these methods are affected by the
high variance of the later histogram bins only if the true depth corresponds to
a later bin. Moreover, it can be seen  from Supplementary
Fig.~\ref{fig:visual_3d_plots}, that using optimal attenuation improves
performance when used in conjunction with a MAP estimator.

\noindent{\bf Disadvantages of MAP Estimation:} The downside of these
estimators is that they require knowledge $\Phi_\text{bkg}$ and
$\Phi_\text{sig}$. While $\Phi_\text{bkg}$ can be estimated easily from data,
estimating $\Phi_\text{sig}$ is difficult to estimate in real-time when the SPAD is
already exposed to strong ambient light.  In comparison, the Coates's estimator
is general and can be applied to any arbitrary flux scenario.

% For the case of a delta waveform of source intensity $\Phi_\text{sig}$ and background $\Phi_\text{bkg}$, the MAP estimator reduces to:
% \[ \widehat d_\mathrm{MAP} = \argmax_d \sum_{i=1}^{d-1} N_i + \frac{1}{\Phi_\text{sig}} \log{\left(\frac{1-e^{-\Phi_\text{bkg}-\Phi_\text{sig}}}{1-e^{-\Phi_\text{bkg}}}\right)} N_d \]
% It can be shown that the source intensity $\Phi_\text{sig}$ needed for depth estimation using the MAP estimator is $O(\sqrt{e^{d \Phi_\text{bkg}}/N})$, compared to $O(\sqrt{e^{n \Phi_\text{bkg}}/N})$ needed using the Coates estimator (The analysis is not very clean, maybe do it in technical report?). Unlike the Coates method, these methods are affected by the high variance of the latter bins only if the actual depth lies there. \\

\section{Simulation details and results}
In this section, we provide details of the Monte Carlo simulations that were
used for the results in the main text. We then provide additional simulation
results illustrating the effect of attenuation. \\

{\bf Details of Monte Carlo Simulation:} We simulate the first photon
measurements using a multinomial distribution as described earlier, for various
background and source conditions. The true depth is selected uniformly at
random from $1$ to $B$, and the simulations are repeated on an average of 200
times. The root-mean-squared depth error (RMSE) is estimated using:
\[
  \mathsf{RMSE} = \sqrt{ \frac{1}{200}\sum_{i=1}^{200}
  \left( \left( \widehat{\tau}_i - \tau^\text{true}_i + \frac{B}{2}\right) \mathsf{mod} B - \frac{B}{2} \right)^2 }
\]
and the relative depth error is calculated as the ratio of the RMSE to the
total depth range:
\[
  \text{relative depth error} = \frac{\mathsf{RMSE}}{B}\times 100.
\]
Here $\tau^\text{true}_i$ denotes the true depth on the $i^\text{th}$
simulation run. It is chosen uniformly randomly from from one of the $B$ bins.
Since the unambiguous depth range wraps around every $B$ bins, we compute the
errors modulo $B$. The addition and subtraction of $B/2$ ensures that the
errors lie in $(-B/2,B/2)$.

\begin{figure}[!ht]
  \centering \includegraphics[width=1\columnwidth]{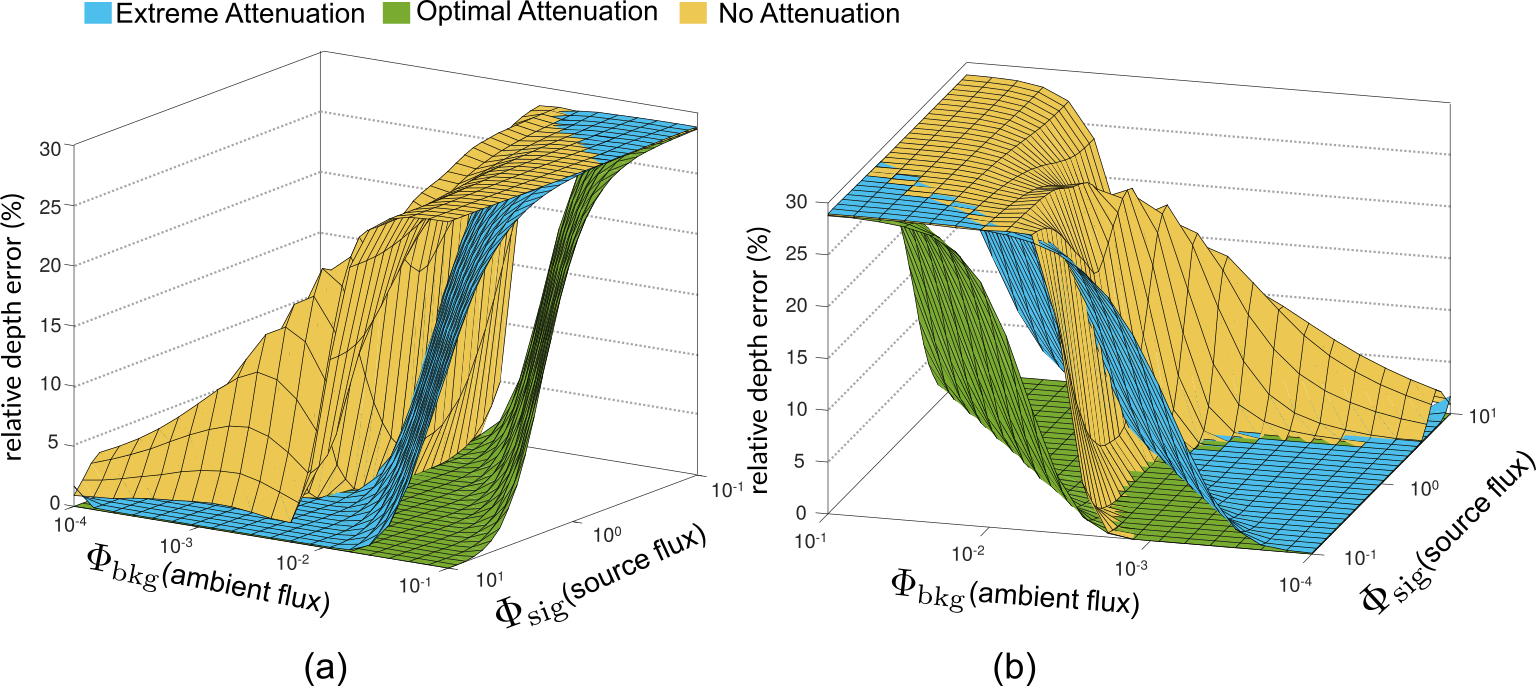}
  \caption{ {\bf Surface plots of relative depth reconstruction errors as a
  function of ambient and source light levels at different attenuation levels.}
  These figures show two different views of surface plots of depth reconstruction
  errors for three different attenuation levels. The optimal attenuation level
  chosen using the $B\Phi_{\text{bkg}} = 1$ performs better than the state-of-the-art
  methods that use extreme attenuation.
  \label{fig:single_pixel_simulation_surface}}
\end{figure}

\subsection{Relative depth error under various signal and background flux conditions}
Supplementary Fig. \ref{fig:single_pixel_simulation_surface}(a) shows the
effect of attenuation on relative depth error, as a 2D function of
$\Phi_{\text{sig}}$ and $\Phi_{\text{bkg}}$ for a wide range of flux
conditions.  It can be seen that with no attenuation, the operable flux range
is limited to extremely low flux conditions. Extreme attenuation extends this
range to intermediate ambient flux levels, but only when a strong enough source
flux level is used. Using optimal attenuation not only provides lower
reconstruction errors at high ambient flux levels but it also extends the range of
SBR values over which SPAD-based LiDARs can be operated. For some
$(\Phi_{\text{bkg}}, \Phi_{\text{sig}})$ combinations, optimal attenuation
achieves zero depth errors, while extreme attenuation has the
maximum possible error of $30\%.$\footnote{Note that the maximum
error of the Coates's estimator is equal to that of a random estimator, which
will have an error of 30\% using the error metric defined earlier.}

Supplementary Fig.~\ref{fig:single_pixel_simulation_surface}(b) shows the same
surface plot from a different viewing angle. It reveals various intersections
between the three surface plots. Optimal attenuation provides lower errors than
the other two methods for all flux combinations.  The error surface when no
attenuation is used intersects the extreme attenuation surface around the
optimal flux level of $\Phi_{\text{bkg}}$ of $0.001$.
For higher $\Phi_{\text{bkg}}$, using no attenuation is worse than using extreme
attenuation, and the trend is reversed for lower $\Phi_{\text{bkg}}$ values.
This is because when $\Phi_{\text{bkg}} \leq 0.001$, the optimal strategy
is to use no attenuation at all. On the other hand, extreme attenuation reduces the flux
even more to $\Phi_{\text{bkg}} = 5e-4$, with a proportional decrease in signal
flux. Therefore, extreme attenuation incurs a higher error.

Also note that while the optimal attenuation and extreme attenuation curves are
monotonic in $\Phi_{\text{bkg}}$ and $\Phi_{\text{sig}}$, the error surface with
no attenuation has a ridge near the high $\Phi_{\text{sig}}$ values. This is an
artifact of the Coates's estimator which we discuss in the next section.

\subsection{Explanation of anomalous second dip in error curves}
Here we provide an explanation for an anomaly in the single-pixel error curves
that is visible in both simulations and experimental results. When
$\Phi_\text{sig}$ is high, increasing $\Upsilon$ beyond optimal has two
effects: the Coates's estimate of the true depth bin becomes higher (due to
increasing effective $\Phi_\text{sig}$), and the Coates's estimates of the
later bins become noisier (due to pile-up). As $\Upsilon$ increases, the
pile-up due to both $\Phi_\text{bkg}$ and $\Phi_\text{sig}$ increases up to a
point where all photons are recorded at or before the true depth bins. Beyond
this high flux level, the Coates's estimates for all later bins become
indeterminate ($N_i = D_i = 0$), and the Coates's depth estimate corresponding
to the location of the highest ratio is undefined.

This shortcoming of the Coates's estimator can be fixed ignoring these bins
when computing the depth estimate. However, since these later bins do not correspond
to the true depth bins, the error goes down. As $\Upsilon$ is increased
further, the pile-up due to ambient flux increases and starts affecting the
estimates of earlier bins too, including the true depth bin. The number of bins
with non-zero estimates keeps decreasing and the error approaches that of a
random estimator.

Note that other estimators like MAP and Bayes do not suffer from the degeneracy
of Coates's estimator since they do not rely on intensity estimates, and should
have U-shaped error curves.

\subsection{Visualization of depth estimation results using 3D mesh reconstructions}
\begin{figure*}[!ht]
\centering \includegraphics[width=0.9\textwidth]{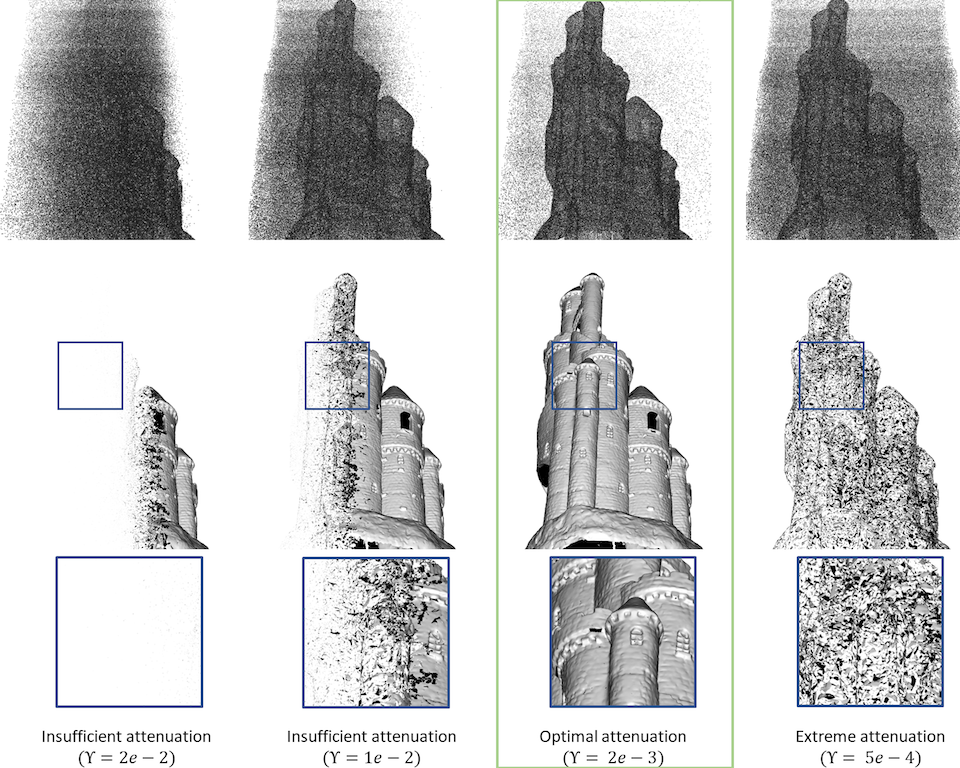}
\caption{ {\bf 3D mesh reconstructions for a castle scene.} (Top row) The raw
point clouds obtained by pixel-wise depth estimation using the MAP estimator.
The haze indicates points with noisy depth estimates. (Bottom row and inset)
The reconstructed surfaces obtained after outlier removal, using ground truth
triangulation. With insufficient attenuation, only the points that are nearby
are estimated correctly, and far away points are totally corrupted. With
extreme attenuation, points at all depths are corrupted uniformly. With optimal
attenuation, most points are estimated correctly, with large depths incurring
slightly more noise due to residual pile-up.
\label{fig:visual_3d_plots}}
\end{figure*}
Supplementary Fig.~\ref{fig:visual_3d_plots} shows 3D mesh reconstructions for
a ``castle'' scene.  For each vertex in the mesh, the true depth was used to
simulate a single SPAD measurement (500 cycles), which was then used to compute
the MAP depth estimate. These formed the raw point cloud. The mesh
triangulation was done after an outlier removal step. These reconstructions
show that nature reconstruction errors is like salt-and-pepper noise, unlike
the Gaussian errors typically seen in other depth imaging methods such as
continuous-wave time-of-flight. Also, it can be seen that as depth increases,
so does the noise (number of outliers). This is because the pile-up effect
increases along depth exponentially. All this suggests that ordinary denoising
methods won't be effective here, and more sophisticated procedures are needed.

\subsection{Improvements from modeling laser pulse shape and SPAD jitter}
\begin{figure*}[!ht]
\centering \includegraphics[width=0.9\textwidth]{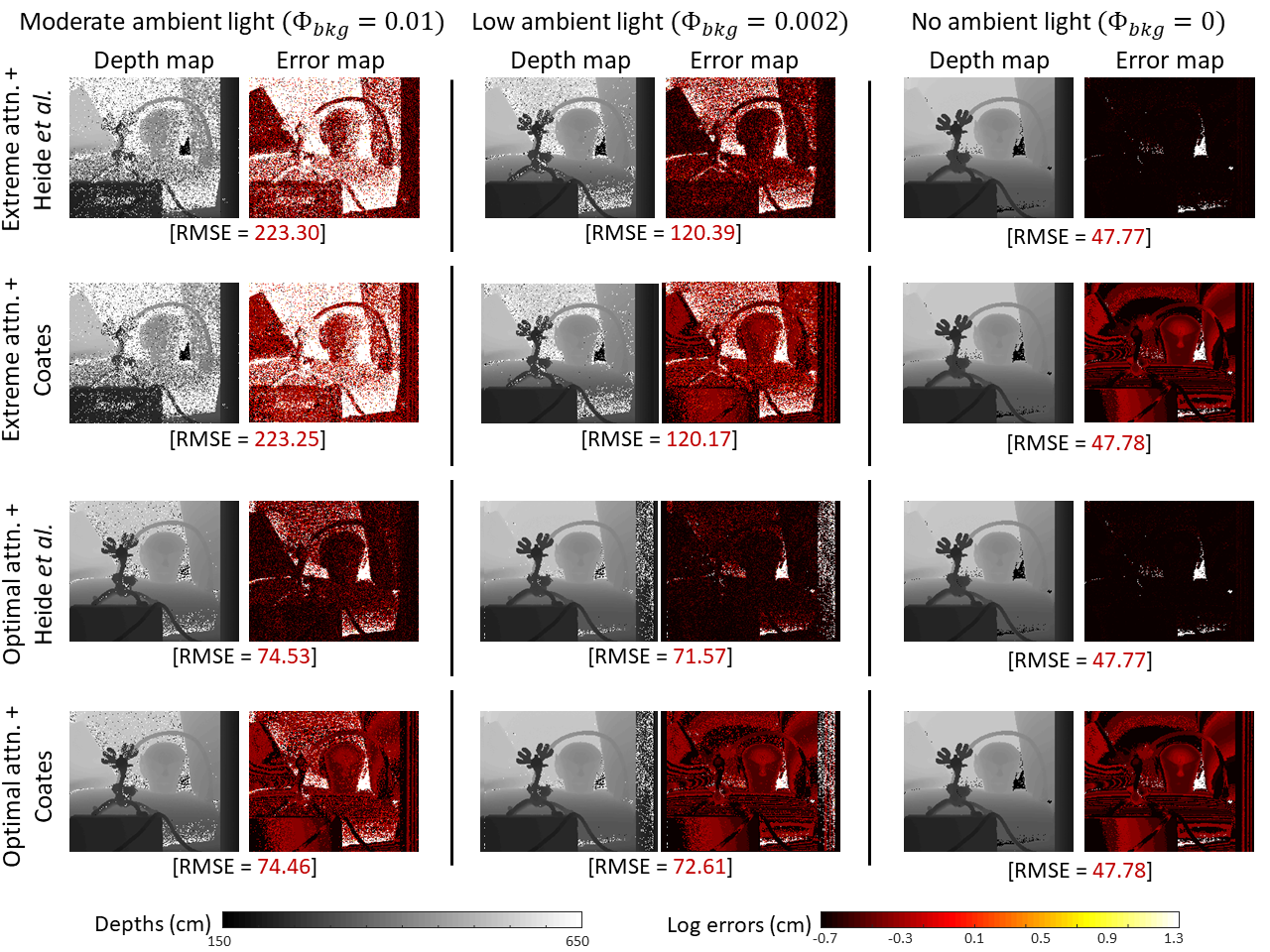}
\caption{ {\bf Effect of modeling laser pulse shape and SPAD jitter, with and without optimal attenuation.} This
figure compares Coates's estimator and Heide \etal's method for the baseline extreme attenuation and the proposed optimal attenuation,
under three levels of ambient light. When the depth errors using Coates's estimator are already low (red pixels in the error maps), Heide \etal's method further reduces error to achieve sub centimeter accuracy (dark red or black pixels). However, for pixels with large errors (white pixels with error $> 10$ cm), Heide \etal's method provides no improvement. The overall RMSE, being dominated by large errors, remains the same. On the other hand, going from extreme attenuation to optimal attenuation reduces depth errors (both visually and in terms of RMSE) for both estimators.
\label{fig:heide}}
\end{figure*}
The depth estimate obtained using the Coates's method
(Eq.~(\ref{eq:coates_corrected_rate_estimate}))
makes the simplifying assumption that the
laser pulse is a perfect Dirac impulse that spans only one histogram bin, even
though our simulation model and experiments use a non-impulse laser pulse shape.
In recent work, Heide \etal \cite{Heide:2018:subpicosecond} 
propose a computational method for pile-up mitigation which
includes explicitly modeling laser pulse shape
non-idealities to improve depth precision.
Suppl. Fig.~\ref{fig:heide} shows simulated depth map reconstructions using the Coates's
estimator and compares them with results obtained using the point-wise
depth estimator of Heide \etal for a range of ambient illumination levels.
Observe that at low ambient light levels, pixels with low error values
with the Coates's estimator appear to be slightly improved in the
depth error maps using the algorithm of
\cite{Heide:2018:subpicosecond}. The method, however, does not improve the
overall RMSE value which is dominated by pixels with very high errors that
stay unchanged.
At high ambient flux levels, pile-up distortion becomes the main source
of depth error and optimal attenuation becomes necessary to obtain good
depth error performance with any depth estimation algorithm. The results
using the total-variation based spatial regularization reconstruction
of \cite{Heide:2018:subpicosecond} did not provide further improvements and
are not shown here. In the next section, we show the effect
of using DNN based methods that use spatial information on the depth
estimation performance for the same simulation scenarios as Suppl.
Fig.~\ref{fig:heide}.

\subsection{Combining attenuation with neural networks-based depth estimation methods}
\begin{figure*}[!ht]
\centering \includegraphics[width=0.9\textwidth]{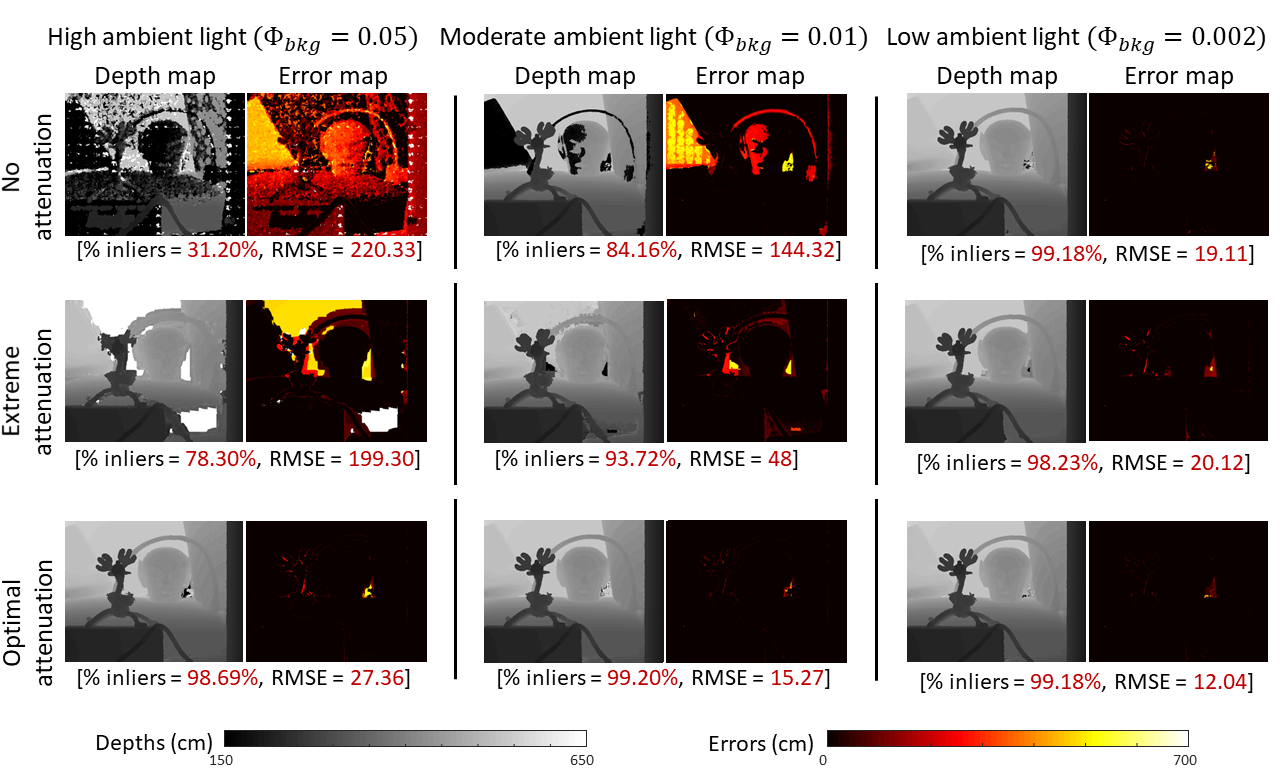}
\caption{ {\bf Effect of attenuation on neural-network based estimator.} This
figure is an extension of Fig.~\ref{fig:simulation_dnn} from the main text,
with three levels of ambient light. Even when ambient light is low, optimal
attenuation leads to an improvement in RMSE compared to extreme and no
attenuation.
\label{fig:neural_network}}
\end{figure*}
\begin{figure*}[!ht]
\centering \includegraphics[width=0.9\textwidth]{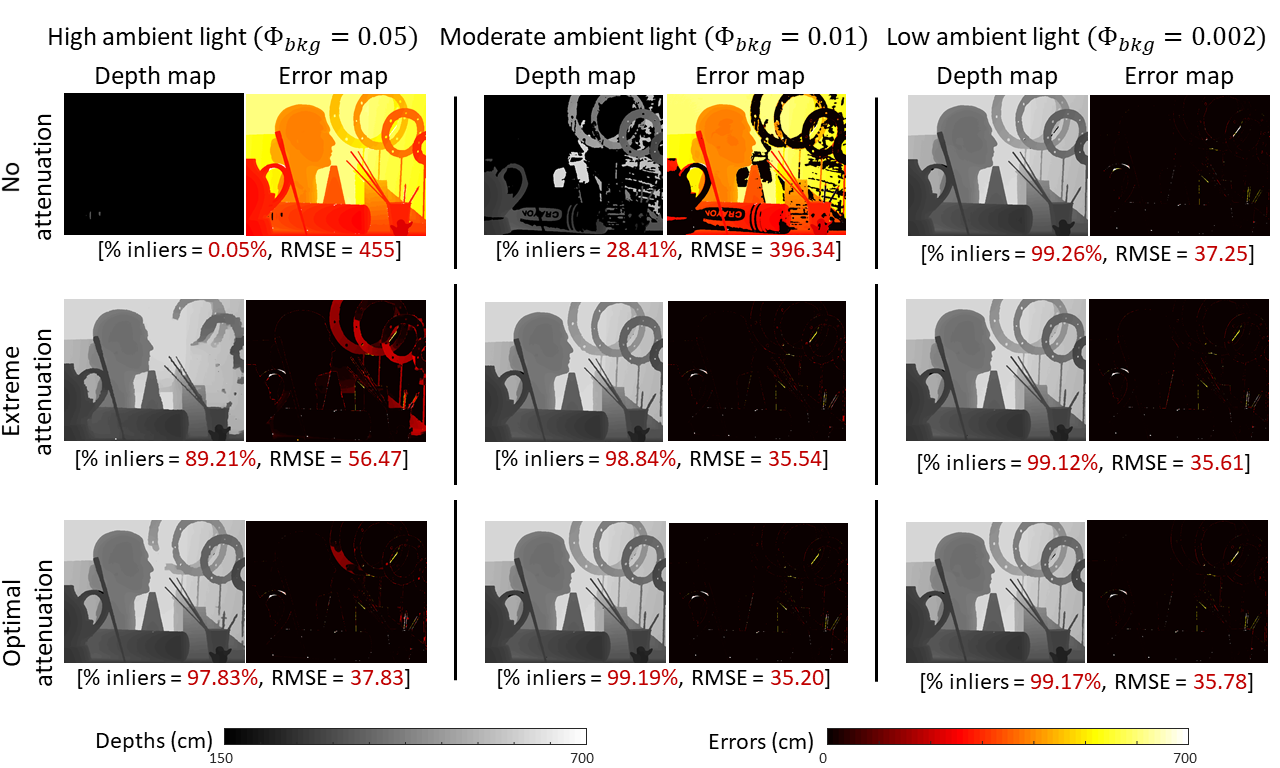}
\caption{ {\bf Effect of attenuation on Rapp and Goyal's method~\cite{Rapp}.} The results follow the same trend as for the neural network.
\label{fig:rapp_and_goyal}}
\end{figure*}

In this section, we provide additional simulation results validating the
improvements obtained from optimal attenuation when used in conjunction with other
state-of-the-art depth reconstruction algorithms. In addition to neural network
based methods, we implemented the method from a paper by  Rapp \emph{et al.}
\cite{Rapp} which exploits spatio-temporal correlations to censor background
photons. Supplementary Fig.~\ref{fig:neural_network} is an extended version of
the Fig.~\ref{fig:simulation_dnn} in the main text and shows reconstruction
results for three different ambient light levels. 

Suppl. Fig.~\ref{fig:rapp_and_goyal} shows the estimated depth maps and errors
obtained using the method from \cite{Rapp} on simulated SPAD measurement data,
for different attenuation and ambient flux levels. These results are similar to
the neural network reconstructions.  For high to moderate ambient flux levels,
the depth estimates appear too noisy to be useful if no attenuation is used.
With extreme attenuation the errors are lower but degrade when the ambient flux
is high. Optimal attenuation provides the lowest RMSE at all ambient flux
levels. 

For the optimal attenuation results shown here, a single attenuation level was
used for the entire scene. The average ambient flux for the whole scene was
used to estimate $\Upsilon^\text{opt}$. This shows that as long as there are
not too many flux variations in the scene, using a single attenuation level is
sufficient to get good performance. For challenging scenes with large albedo or
lighting variations, a single level may not be sufficient and it may become
necessary to use a patch-based or pixel-based adaptive attenuation. This strategy
is discussed in the next section.

\section{Ambient-adaptive attenuation}
This section describes an algorithm for implementing the idea of optimal
attenuation in practice. The only variable in the expression for optimal
$\Upsilon$ is the background flux $\Phi_{\text{bkg}}$, which can be estimated
separately, prior to beginning the depth measurements. For estimating
$\Phi_{\text{bkg}}$, the laser is turned off, and $N'$ SPAD cycles are
acquired. Since the background flux $\Phi_{\text{bkg}}$ is assumed to be
constant, there is only one unknown parameter, and it can be estimated from the
acquired histogram $(N_1', N_2', ..., N_B')$ using the MLE (Step 3 of
Algorithm~\ref{alg:ndfilter}).  Moreover, as mentioned in the main text, our
method is quite robust to the choice of $\Upsilon$, which means that our
estimate of $\Phi_{\text{bkg}}$ does not need to be very accurate. Therefore,
we can set $N'$ to be as low as 20--30 cycles, which causes negligible
increase in acquisition time.

\begin{algorithm}
\caption{Adaptive ND-attenuation\label{alg:ndfilter}}
\begin{myenum}
 \item Focus the laser source and SPAD detector at a given scene point.
 \item With the laser power set to zero, acquire a histogram of photon counts
   $(N'_1, N'_2, \ldots, N'_{B+1})$ over $N'$ laser cycles.
 \item Estimate the background flux level using:
   \[
     \widehat \Phi_\text{bkg} = \ln \left( 
     \frac{\sum_{i=1}^B iN'_i + B N'_{B+1}}{\sum_{i=1}^{B+1}i N_i'-N'}
     \right).
   \]
 \item Set the ND-attenuation fraction to $\nicefrac{1}{B\widehat\Phi_\text{bkg}}$.
 \item Set the laser power to the maximum available level and acquire a
   histogram of photon counts $(N_1, N_2, \ldots, N_{B+1})$ over $N$ laser
   cycles.
 \item Estimate the photon flux waveform using the Coates's correction
   Equation~(\ref{eq:coates_corrected_rate_estimate}), and scene depth using
   Equation~(\ref{eq:time_of_arrival_estimator}).
 \item Repeat for all scene points.
\end{myenum}
\end{algorithm}

\section{Dependence of reconstruction errors on true depth value}
\begin{figure*}[!ht]
\centering \includegraphics[width=1\textwidth]{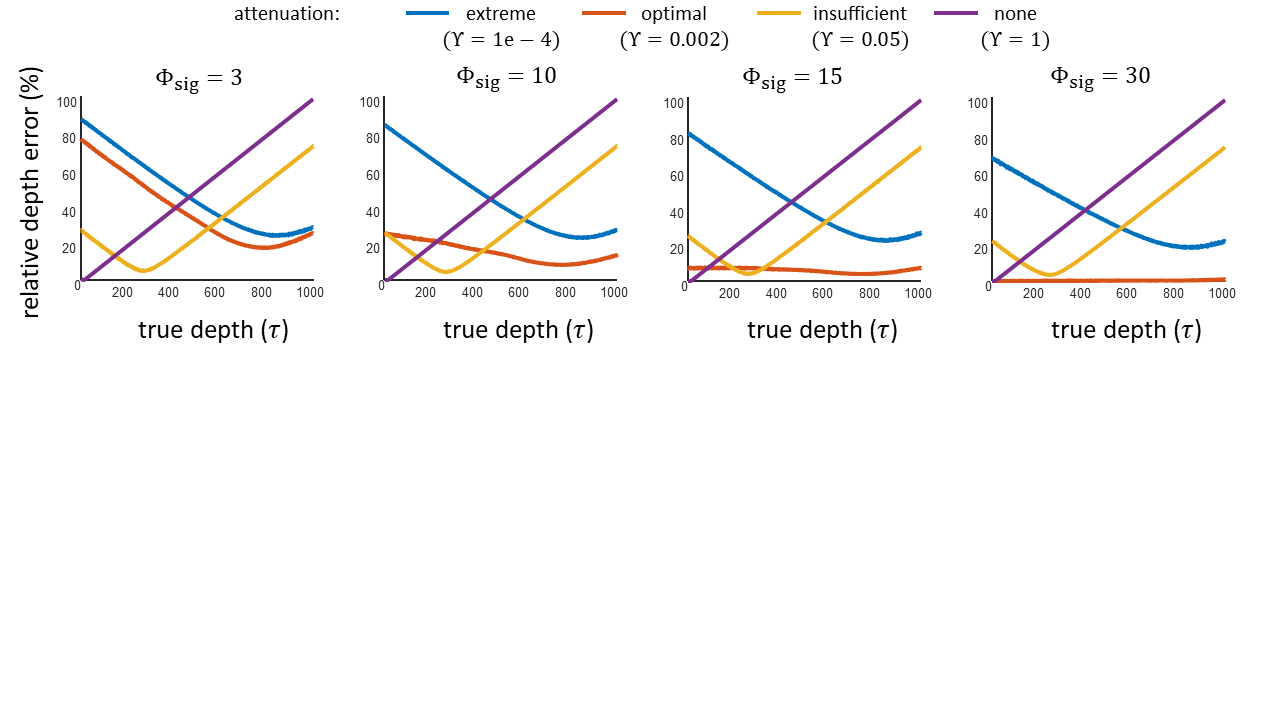}
\vspace{-1.8in}
\caption{ {\bf Effect of attenuation on error vs true depth curve.} For
extreme, insufficient and no attenuation, the error curves are not only high on
average, but also highly non-uniform (either decreasing or increasing with
depth). In contrast, the optimal attenuation curve is both low on average, and
relatively uniform across depth.  
\label{fig:error_vs_depth}}
\end{figure*}
In this section, we study the effect of true depth on depth estimation errors.
Due to the non-linear nature of the image formation model, as well as the
non-linear estimators used to rectify pile-up, the estimation error shows some
non-linear variations as a function of  the true depth.

Suppl. Fig.~\ref{fig:error_vs_depth} compares depth error curves across
various attenuation levels, for a several signal values. The first observation
is that optimal attenuation has a lower error, on average, than extreme and no
attenuation. The error curve for optimal attenuation lies below the other
curves for most values of true depth (except for very low signal). Therefore,
not only does optimal attenuation minimize the average error, it makes the
error curve more uniform across all values of the true depth.

\clearpage
\section{Additional Experimental Results}
\begin{figure*}[!ht]
\centering \includegraphics[width=0.8\textwidth]{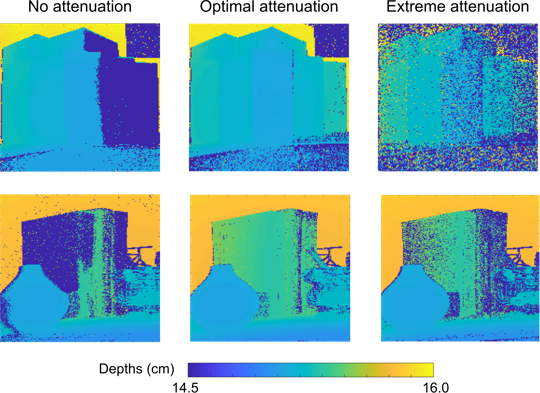}
\caption{ {\bf Depth estimation with different attenuation factors.} (Top row)
Depth maps for a staircase scene, with a brightly lit right half, and shadow on
the left half. With no attenuation, the right half is completely corrupted with
noise due to strong pile-up. (Bottom row) A challenging tabletop scene with
large albedo and depth variations. The optimal attenuation method still gives a
reasonably good reconstruction, and is significantly better than either no
attenuation or extreme attenuation.   
\label{fig:single_pixel_experiment_point_scan}}
\end{figure*}

\begin{figure}
\centering \includegraphics[width=0.5\columnwidth]{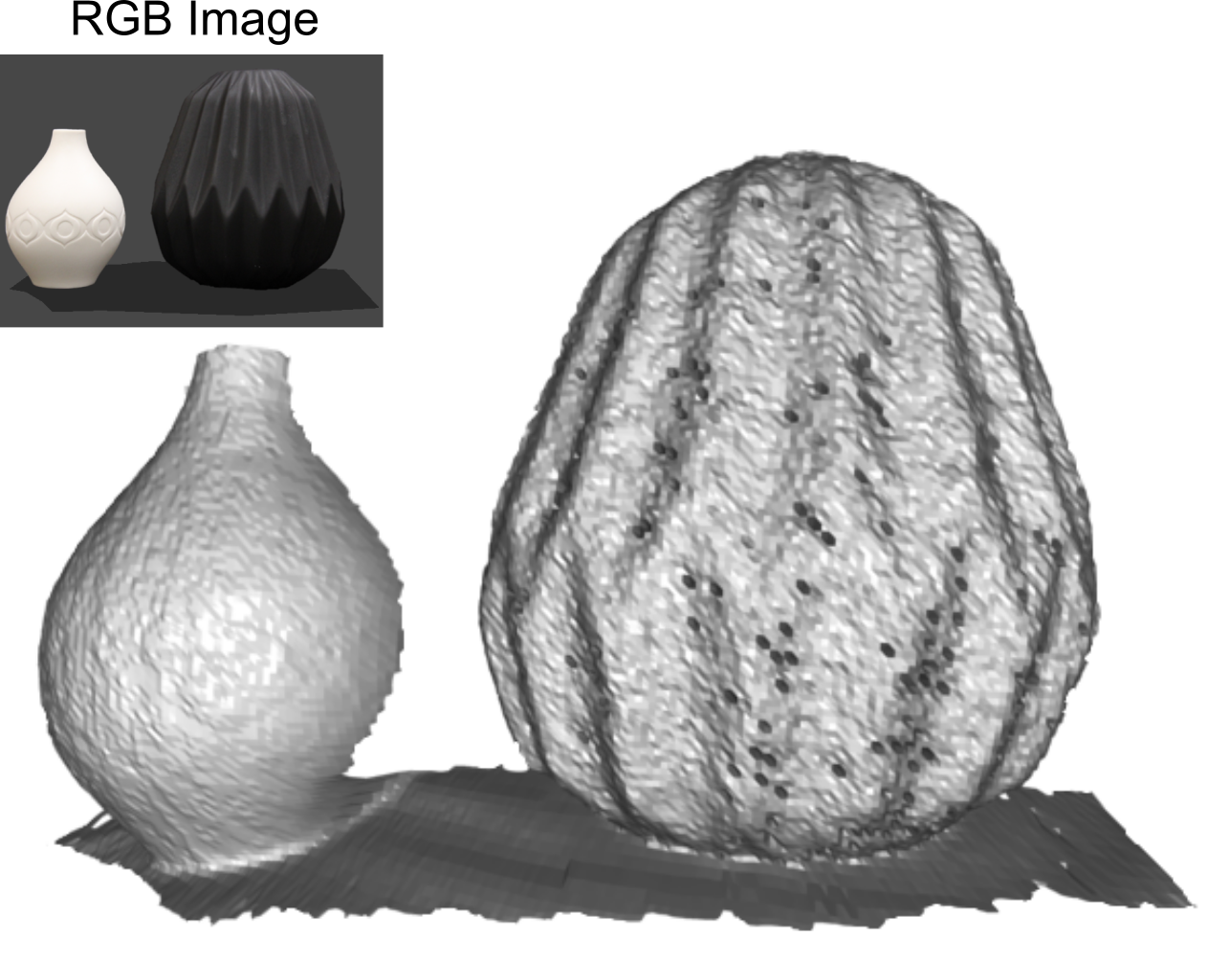}
\caption{ 
{\bf Reconstructing extremely dark objects.} Our method works for scenes with a
large dynamic range of flux conditions, like this scene with an extremely dark
black vase placed next to a white vase. $\Phi_\text{sig}$ was 10 times higher
for the white vase. This scan was done with negligible ambient light ($<10$
lux).}
\end{figure}

\end{document}